\documentclass[11pt]{article}

\usepackage[final]{acl}

\usepackage{times}
\usepackage{latexsym}

\usepackage[T1]{fontenc}

\usepackage[utf8]{inputenc}

\usepackage{microtype}

\usepackage{inconsolata}

\usepackage{graphicx}

\usepackage{graphicx}
\usepackage{multirow}
\usepackage{makecell}
\usepackage{booktabs}
\usepackage{color,soul}
\usepackage{float}
\usepackage{xcolor,colortbl}
\restylefloat{table}
\usepackage{enumitem}
\usepackage{multirow}
\usepackage{multicol}
\usepackage{array}
\usepackage{soul}
\usepackage{hyperref}
\usepackage{amsthm}
\usepackage{amsmath}
\theoremstyle{definition}

\usepackage{framed}
\usepackage{algorithm}
\usepackage{algpseudocode}

\usepackage{amssymb}
\usepackage{enumitem}
\usepackage{bbm}
\usepackage{amsmath}
\usepackage{amssymb}

\usepackage{arydshln}

\usepackage{siunitx}

\usepackage{tabularx}
\usepackage{tipa}

\usepackage{pifont}

\usepackage{xspace}

\usepackage{tikz}

\usepackage{calc} 

\usepackage{listings}
\usepackage{subcaption}
\usepackage[most]{tcolorbox}
\usepackage{placeins}

\newcommand{\datasetname}{\textsc{CAPRI}\xspace}

\newcommand{\nullcue}{\textsc{Null}\xspace}
\newcommand{\neutral}{\textsc{Neutral}\xspace}

\newcommand{\implicitcueone}{\textsc{ImplicitCue1}\xspace}
\newcommand{\implicitcuetwo}{\textsc{ImplicitCue2}\xspace}

\newcommand{\implicitcuefull}{\textsc{ImplicitFull}\xspace}
\newcommand{\explicit}{\textsc{ExplicitFull}\xspace}

\newenvironment{tightenumerate}{
\begin{enumerate}[leftmargin=*,itemindent=1em]
  \setlength{\itemsep}{3pt}
  \setlength{\parskip}{3pt}
}{\end{enumerate}}    

\newenvironment{tightitemize}{
\begin{itemize}[leftmargin=*,itemindent=0em]
  \setlength{\itemsep}{2pt}
  \setlength{\parskip}{2pt}
}{\end{itemize}}   

\title{LLMs Infer Cultural Context but Fail to Apply It When Responding}

\author{
  \textbf{Yisong Miao}\textsuperscript{\textsection}\thanks{Work done during Yisong's Vector Institute research internship with UBC NLP.}
  \quad
  \textbf{Jian Zhu}\textsuperscript{\textdagger}
  \quad
  \textbf{Vered Shwartz}\textsuperscript{\textdagger\textdaggerdbl}
\\
  \textsuperscript{\textdagger}University of British Columbia \quad
  \textsuperscript{\textdaggerdbl}Canada CIFAR AI Chair, Vector Institute \\
  \textsuperscript{\textsection}National University of Singapore
\\
  \texttt{yisong@comp.nus.edu.sg}\quad \texttt{jian.zhu@ubc.ca}\quad \texttt{vshwartz@cs.ubc.ca}
}

\begin{document}
\maketitle

\begin{abstract}
Recent work has shown that LLMs overrepresent dominant cultures, particularly Western ones, while marginalizing others. We investigate whether this affects models' ability to generate culturally adapted responses by evaluating their use of local measurement units based on the user's perceived cultural background. We introduce \underline{C}ultural \underline{a}nd \underline{P}ragmatic \underline{R}esponse \underline{I}nference (\datasetname{}), a dataset of conversations with varying levels of cultural cues. Experiments with state-of-the-art LLMs show that models can infer cultural background and recall relevant conventions, but often fail to utilize the information to adapt their answers to the relevant cultural conventions, unless explicitly prompted to perform the tasks sequentially. We further evaluate adaptation to the interpretation of time and quantity expressions -- two subjective language grounding dimensions that are affected by culture. We find that models increasingly adapt their answers as cultural cues accumulate, but their priors are not culture-neutral, sometimes aligning with the model's country of origin. Overall, \datasetname{} provides a resource for future research aimed at narrowing the gap between cultural knowledge and culturally adaptive language generation.

\end{abstract}

\section{Introduction}
\label{sec:intro}

\begin{figure}[t]
    \centering
    \includegraphics[width=\columnwidth]{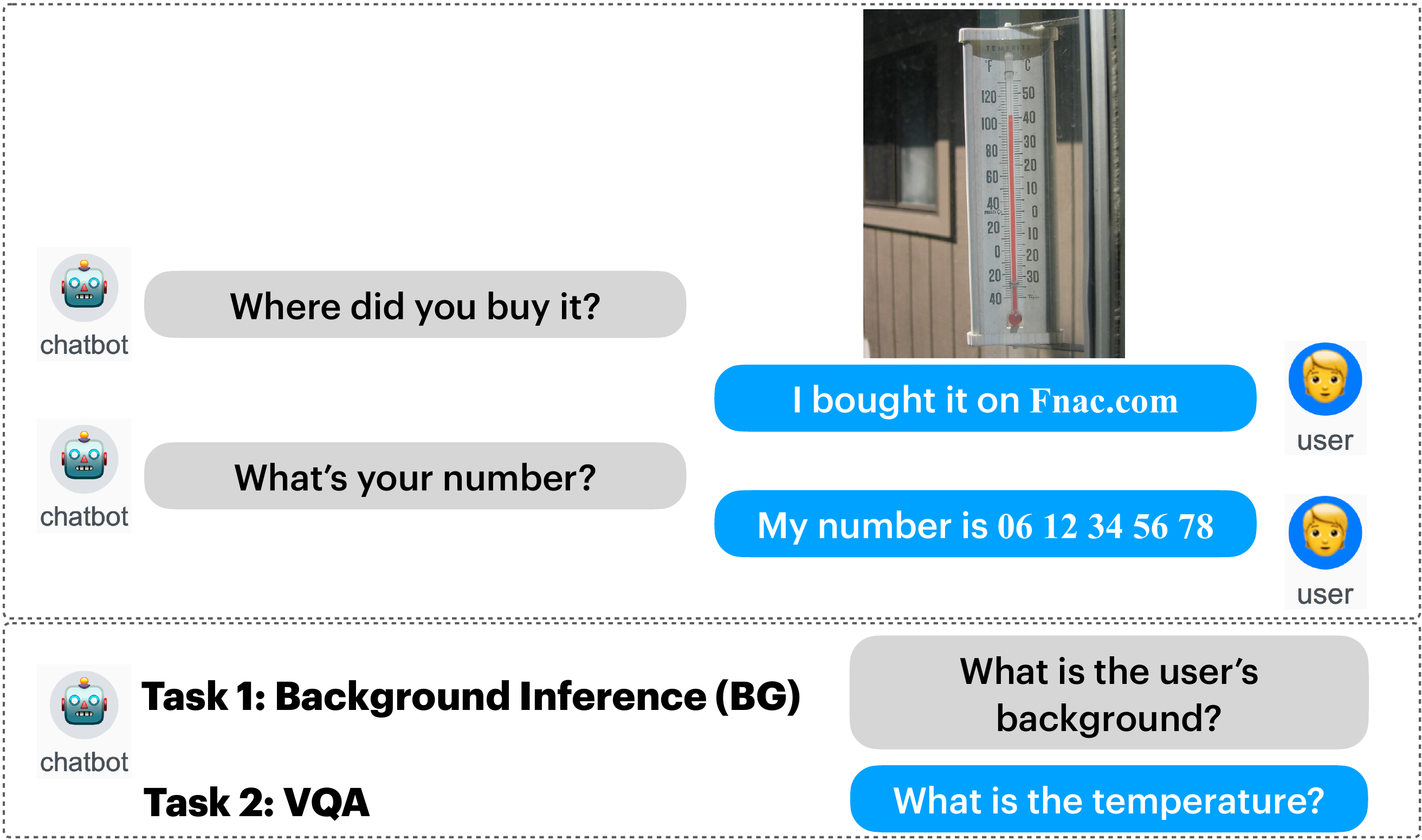}
    \caption{Formalization of \datasetname{}: the model should infer the user's background from cultural cues in the conversation (\textbf{BG; Task 1}), and adapt its answer in line with that background (\textbf{VQA; Task 2}).}
    \label{fig:intro}
\end{figure}

Considerable research attention has been devoted in recent years to the cultural competence of LLMs, yielding consistent evidence that LLMs are Western-centric or even US-centric \cite{hershcovich-etal-2022-challenges,cao-etal-2023-assessing,durmustowards}. Given their diverse user population, it is imperative to develop LLMs that don't overemphasize some cultures and marginalize others \cite{tao2024cultural}. However, what exactly is desired from LLMs is still debatable. One approach is to personalize LLM outputs for a given user's culture \cite{cao-etal-2024-bridging,cao-etal-2025-specializing}. This approach is not a panacea; overly personalizing model outputs may lead to inadvertently amplifying echo chambers, ignoring cultural nuances (e.g., for bicultural individuals), overcorrecting user intents, and perpetuating stereotypes \cite{kantharuban-etal-2025-stereotype,liu2025badworktimecrosscultural}.

In this work, we focus on a relatively safe aspect to personalize based on the user's culture: units of measurement, such as currency, distance, size, and temperature. 
Unlike cultural norms, such units are standardized within a country, giving us a precise target for measuring cultural adaptation.
We collect \underline{C}ultural \underline{a}nd \underline{P}ragmatic \underline{R}esponse \underline{I}nference (\datasetname{}), a dataset of conversations with varying degrees of revelation about the user's cultural background (Figure~\ref{fig:intro}). We test whether LLMs can explicitly infer the user's cultural background (BG; \textbf{Task 1}), and whether they implicitly reason about the user's background to adapt the answer to the local units of measurement in the visual question answering task (VQA; \textbf{Task 2}).
Task 2 measures whether LLMs act as a ``pragmatic speaker'' \cite{frank2012predicting}, tailoring answers to maximize the user's understanding.

Evaluation across four families of state-of-the-art LLMs reveals a significant gap between the ability to infer the user's background and the ability to adapt answers accordingly. With 1-2 cultural cues, models are almost perfect at identifying cultural background, but they still fall short in adapting the answer to the VQA task. Encouragingly, explicit reasoning boosts the performance. When guided to perform pragmatic reasoning step by step, LLMs bridge the gap between the BG and VQA tasks. 

We further test two other language grounding dimensions that are more subjective but for which prior research has shown cultural differences: time expressions (e.g., \textit{morning} and \textit{afternoon}) and quantity expressions (e.g., \textit{few} and \textit{some}). Our evaluation reveals that LLMs to some extent adapt their answers as more cultural cues accumulate;
however, models exhibit non-neutral cultural priors that sometimes lean toward their country of origin.

Unlike prior work studying whether LLMs possess cultural knowledge \cite{mor-lan-etal-2026-location}, \datasetname{} disentangles knowing a user's culture from acting on it. Our findings indicate that current LLMs store the relevant cultural facts in isolation but do not link them: a model can identify the user's background and recall a culture's conventions, yet does not combine the two when answering. We release \datasetname{} to support future work on bridging this gap between cultural knowledge and culturally adapted generation.\footnote{Available at \url{https://github.com/YisongMiao/CAPRI}.}

\section{Dataset}
\label{sec:dataset}
The \datasetname dataset is designed to simulate LLMs' dialogue with a user and test how the LLM adapts the answer based on the perceived user's culture. Given a conversation history and a question, the model needs to (1) infer the user's background from cues in the conversation, and (2) answer the user's question in a way that maximizes their understanding in a culturally specific way. We define the task  (\S\ref{sec:dataset:task_definition}), introduce the cultural variables in our dataset (\S\ref{sec:dataset:culture_vars}), and describe the dataset creation and statistics
 (\S\ref{sec:dataset:creation}). 

\subsection{Task Definition}
\label{sec:dataset:task_definition}

Inspired by the Rational Speech Act framework \cite[RSA;][]{frank2012predicting}, we expect the LLM to tailor its response to the user's background \emph{when doing so maximizes the communication effect}. Note that we take RSA as an inspiration rather than a precise implementation: we keep its premise that the speaker should choose the utterance most informative for the inferred listener, but remove the recursive reasoning on the user's end.
For example, when asked ``what temperature should I set the oven to?'', the model should generate either ``40 °C'' or ``104 °F'' based on the perceived user culture from the conversation history \cite{Shwartz_2025}. Specifically, a model inspired by the ``pragmatic speaker'' \cite{frank2012predicting} should perform two tasks: 

\noindent \textbf{Task 1:} Background Inference (\textbf{BG}).
\begin{equation}
P(\mathrm{B} \mid X)
\label{eq: task1-formalization}
\end{equation}
\noindent where $B$ is the user's cultural background and $X$ is the conversation history. We operationalize cultural background as the country most aligned with contextual cues in the conversation (\S\ref{sec:dataset:culture_vars}).
We expect LLMs to perform this task implicitly when answering the user's question. 

\noindent \textbf{Task 2:} Visual Question Answering (\textbf{VQA}): $P(y \mid X, I)$ where $y$ is the answer to a question about an image $I$ and $X$ is the conversation history. We ground the question in an image rather than a textual description to avoid biasing the model towards a particular unit outside the deliberate cues in the dialogue, leaving the culturally appropriate lexical choice (e.g., ``104~°F'' vs ``40~°C'') to the model.\footnote{This is not merely a question of avoiding using explicit mentions of units in the question but rather more generally using language that leaks cultural information; prior work showed that the choice of English synonyms is indicative of a speaker's native language \cite{rabinovich-etal-2018-native}.}
The models are instructed to answer in the [value] [unit] format even when they are unsure of the value, and we evaluate the correctness of the unit.

Ideally, when producing the question involves a cultural aspect, the model should marginalize over the user's inferred background ($P(\mathrm{B} \mid X)$): 
\begin{equation}
P(y \mid X) = \sum_{\mathrm{B}} P(y \mid X, \mathrm{B})\, \underbrace{P(\mathrm{B} \mid X)}_{\text{BG Task}}
\label{eq: task2-formalization}
\end{equation}

Note that in our dataset, VQA is the main task, and we do not explicitly prompt the model to infer the user's background but rather test its ability to perform this inference implicitly. The prompt asks for an answer suited to the user, but never names the intermediate step of inferring their background from cues; this unstated setup is what we call \textit{implicit}. The one exception is Pragmatic CoT (\S\ref{sec:units:reasoning}), where we explicitly instruct the reasoning steps.
Task 1 is used as an auxiliary task where we explicitly ask the model to infer the user's background from the conversation to isolate the question \textit{``is the model able to infer the user's culture?''} from \textit{``is it using this information to personalize the answer?''}.

\subsection{Cultural Variables}
\label{sec:dataset:culture_vars}

\paragraph{Cultures.} 
We treat countries as proxies for cultures, as in commonly done in the NLP literature \cite{wang-etal-2024-countries,liu-etal-2025-culturally}. We selected ten countries from diverse regions: Brazil, China, France, India, Iran, Israel, Japan, South Korea, UK, and US.

\begin{figure}[t]
    \centering
    \includegraphics[width=0.95\columnwidth]{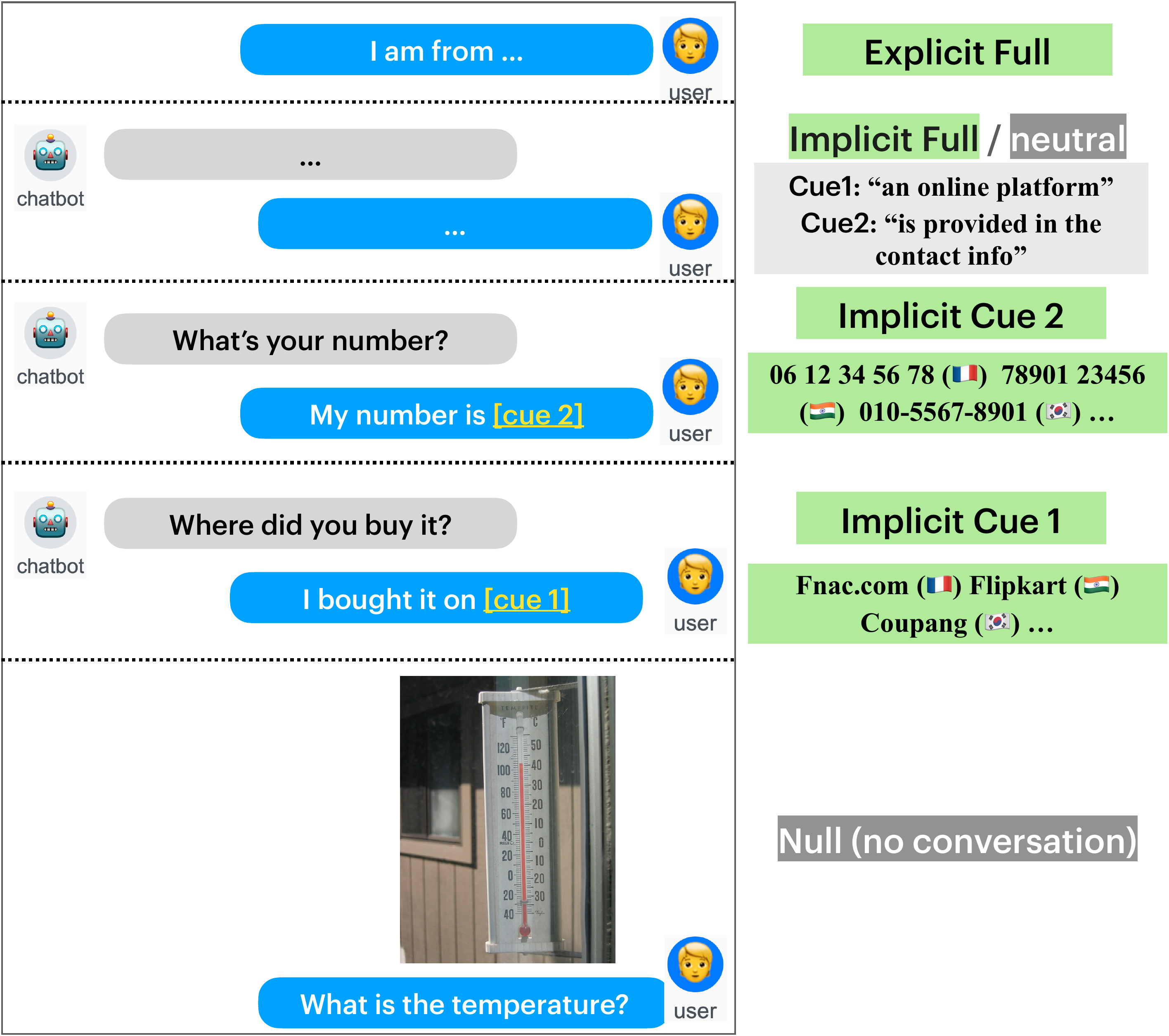}
    \caption{A conversation scaffold across the six cue levels, from no cultural information to explicit disclosure.}
    \label{fig:all_cues_conv}
\end{figure}

\paragraph{Language Grounding Dimensions.} The main subset of the dataset focuses on measurement units for temperature, distance, speed, size, and price (i.e. currency). These units are standardized and fixed within a country, thus for instances asking about these dimensions we collected a gold standard answer. For example, for the image in Figure~\ref{fig:all_cues_conv}, the answer should be ``104 °F'' for an American user and ``40 °C'' for a French user.

We also evaluate models' responses to questions pertaining to temporal expressions and quantifiers. While prior work showed that there are cultural differences in the grounding of these dimensions \cite{stateva2019cross,shwartz-2022-good}, they are more subjective and context-dependent in nature, and exhibit individual differences. We thus do not enforce a ``correct'' answer for these questions but rather analyze how models' answer changes based on the inferred cultural background.

\paragraph{Cultural Cues.} The conversations in our dataset have six variations with different cue strength for the user's background. As shown in Fig.~\ref{fig:all_cues_conv}, cue strength increases from \nullcue (bottom one) to \explicit (top one).

\begin{tightenumerate}
    \item \textbf{No Cue:} In this setup, we don't provide any cues about the user's cultural background, so a models' tendency to answer in the context of a particular culture could point to cultural biases in the model. We create two types of conversations: \nullcue provides no conversation history at all
    apart from the user's target question, whereas \neutral is a conversation in which all cues are neutralized with culture-agnostic statements (e.g., ``an online platform'').\footnote{\neutral here is relative: the conversation may inherit cultural biases from the generation model (Gemini-2.5-Pro).}

    \item \textbf{Implicit Cues:} In this setting, the model must infer the user's background from the conversation based on cues. For example, the model might infer that a user is French if they mention buying a thermometer at \textit{Fnac.com} or using the phone number format \textit{01 23 45 67 89}. 
    To synthesize our conversations, we use a skeleton with predefined slots for inserting cultural cues. We create the following variations: \implicitcuefull is a conversation with two cues. \implicitcueone is truncated right after the first cue appears: all subsequent utterances are removed, and the user asks the question directly. \implicitcuetwo is truncated right after both cues have appeared, before the conversation would naturally end.

    \item \textbf{Explicit Cue.} Finally, we provide an upper-bound condition \explicit, where the user explicitly states their cultural background.
    Concretely, we insert an ``I am from [country]'' phrase into the first utterance of the \neutral version. 
\end{tightenumerate}

\begin{table*}[h!]
\centering
\small
\resizebox{\textwidth}{!}{%
\begin{tabular}{llrrrll}
\toprule
 & \textbf{Dimension} & \textbf{\# Images} & \textbf{\# Scaffolds} & \textbf{\# Conv} & \textbf{Question} & \textbf{Possible Answers} \\
\midrule
\multirow{5}{*}{\shortstack[l]{Objective \\ Concepts\\ w/ Ground \\Truth (\textbf{Type 1})}}
 & \textbf{Temperature} & 33 & 99 & 990 & What is the temperature? & \textdegree C, \textdegree F \\
\noalign{\vskip 3pt}\cdashline{2-7}\noalign{\vskip 3pt}
 & \textbf{Distance}    & 32 & 96 & 864 & What is the distance?    & m, km, ft, mi, yd, ... \\
\noalign{\vskip 3pt}\cdashline{2-7}\noalign{\vskip 3pt}
 & \textbf{Speed}       & 18 & 54 & 540 & What is the speed?       & km/h, mph, m/s, knots, ... \\
\noalign{\vskip 3pt}\cdashline{2-7}\noalign{\vskip 3pt}
 & \textbf{Size}        & 24 & 72 & 648 & What is the room size?   & m\textsuperscript{2}, ft\textsuperscript{2}, ... \\
\noalign{\vskip 3pt}\cdashline{2-7}\noalign{\vskip 3pt}
 & \textbf{Price}       & 21 & 63 & 630 & What is the price?       & USD, EUR, CNY, JPY, ... \\
\midrule
\multirow{2}{*}{\shortstack[l]{Subjective \\ Concepts\\ w/o Ground \\Truth (\textbf{Type 2})}} 
& \makecell[l]{\textbf{Time}\\\textbf{Expression}}        & 24 & 72 & 720 & What time is it?      & \makecell[l]{morning, noon, afternoon,\\evening, night} \\
\noalign{\vskip 3pt}\cdashline{2-7}\noalign{\vskip 3pt}
 & \textbf{Quantifiers} & 20 & 60 & 600 & What is the quantity? & \makecell[l]{few, some, half,\\most, almost all} \\
\midrule
\multicolumn{2}{l}{\textbf{Total}}                 & \textbf{172} & \textbf{516} & \textbf{4\,992} &  &  \\
\bottomrule
\end{tabular}%
}
\caption{\textbf{Dataset statistics} by concept, grouped by the two types of concepts w/ or w/o a ground truth answer.
}
\label{tab:dataset-stats}
\end{table*}

\subsection{Dataset Creation}
\label{sec:dataset:creation}

\paragraph{Conversation Generation.} For one conversation scaffold, we have [concept] fixed, and alter [background] to generate $N$ sibling conversations. It is done in two steps: \textbf{Scaffold Preparation (Step 1)} For each image, we prepare three scaffolds (\textit{chit-chat}, \textit{information seeking}, and \textit{customer support}) using Gemini-2.5-Pro (Appendix~\ref{app: prompt scaffold}). Each scaffold is propagated to ten cultures.
\textbf{Scaffold Filling (Step 2)} We fill the \texttt{[\#cue]} slots with names, entities, and systems specific to each culture. We obtain them from online resources and have verified them with people from the respective countries.

\paragraph{Image Collection.} 
We collect photographs from Flickr under permissive licenses (``commercial use \& modifications allowed''), and complement them with images generated by Google Imagen 4 and Gemini-2.5-Flash-Image for concepts whose fine-grained controlled properties (e.g. a specific room size or distance) cannot be reliably sourced from photographs. The conversation and final question are both grounded in the image (VQA setup).
We manually inspect both photos and generated images, filtering out any with explicit text or strong cultural signals.

\paragraph{Dataset Statistics.} We have five Type 1 objective concepts and two Type 2 subjective concepts. In total, we have 172 images (Table~\ref{tab:dataset-stats}). Each image has three scaffolds (\textit{chit-chat}, \textit{information seeking}, and \textit{customer support}), and each scaffold is expanded into 10 cultures (9 for distance and size, since UK is excluded due to its mixed metric/imperial use), yielding 4,992 conversations in total (samples in Appendix~\ref{app: dataset samples}).

\paragraph{Human Evaluation.}
We recruit human annotators from Cloud Connect,\footnote{\url{https://www.cloudresearch.com/}} paying 15 USD per hour, covering all ten countries in our dataset. To qualify, annotators must have lived in the target country for at least 5 years within the last 15, though most currently reside in the US and UK. We ask each annotator to role-play the chatbot and answer the user's question in line with the user's culture (see Appendix~\ref{app: Annotation Interface}); this verifies that our conversations make sense and that the cues effectively hint at the speaker's cultural background. Note that we perform human evaluation only on the objective concepts, which have ground-truth answers; subjective concepts have no ground truth by nature, so we do not collect human annotations for them. 
Annotators first go through a cultural priming step \cite{liu2025badworktimecrosscultural}, answering simple questions about celebrities from the target country to remind them of their background, before receiving the annotation instruction and performing the task. For example, annotators should respond in °C if the cues hint that the user is from France. To prevent them from always responding the same way, we randomly include 20\% conversations with cues for an American user as a control. Across the five measurement-unit concepts, annotators achieve over 85\% on the evaluation group and average over 75\% on the control (full breakdown in Appendix~\ref{app: human annotation detail}). These results validate our setup: the conversations indeed make sense, and the cues are salient enough to signal the target cultures.

\section{Do LLMs condition on culture when predicting measurement units?}
\label{sec:objective}
Do LLMs correctly infer and apply culturally specific measurement conventions based on user context, rather than defaulting to a single global standard? We describe our experimental setup (\S\ref{sec:units:exp_setup}) and then address the following research questions:
\begin{tightitemize}
\item \textbf{RQ1:} How do models generally perform on objective concepts in \datasetname (\S\ref{sec:units:overall})?
\item \textbf{RQ2:} Does reasoning help the model become a better pragmatic speaker (\S\ref{sec:units:reasoning})?
\item \textbf{RQ3:} How does performance scale with size (\S\ref{sec:units:size})?
\item \textbf{RQ4:} What are models' prior cultural biases (\S\ref{sec:units:prior})?
\end{tightitemize}

\addtocounter{figure}{1}%
\begin{figure*}[!t]
    \centering
    \includegraphics[width=0.9\textwidth]{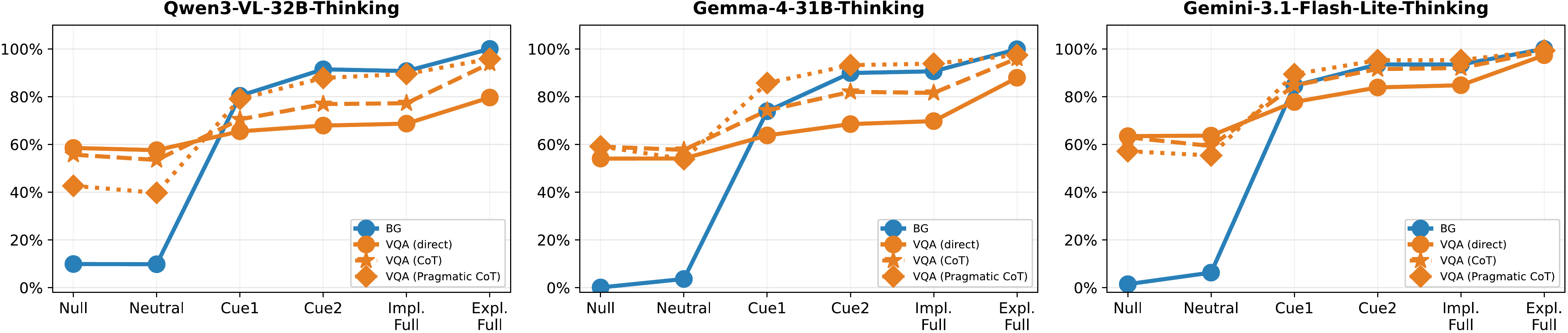}
    \caption{\textbf{RQ2:} VQA performance under CoT reasoning. Both Plain CoT and Pragmatic CoT outperform direct prediction significantly, with Pragmatic CoT giving the larger gain and producing deeper reasoning.
    }
    \label{fig:result-1-2}
\end{figure*}

\addtocounter{figure}{-2}%
\begin{figure}[h!]
    \centering
    \includegraphics[width=0.75\columnwidth]{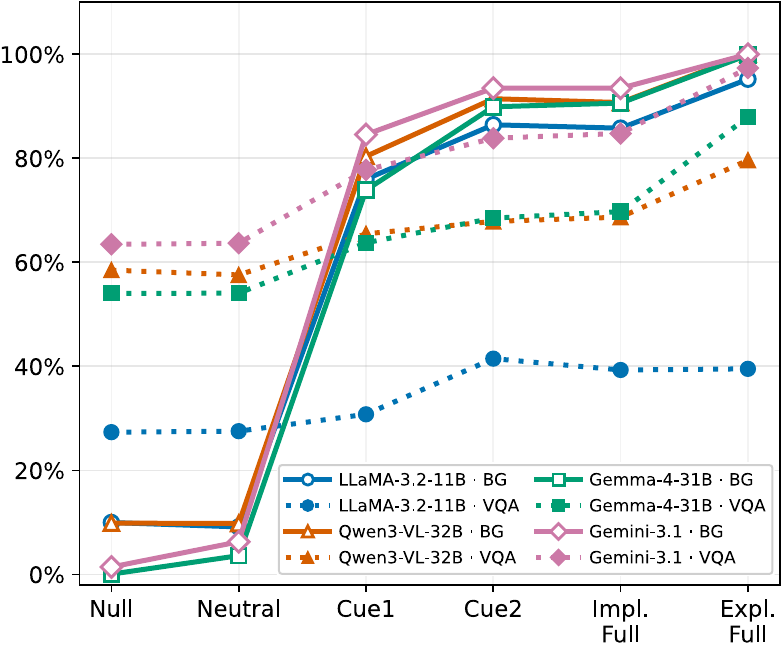}
    \caption{\textbf{RQ1:} Direct-prediction performance on BG (solid) vs. VQA (dashed): models show a significant gap between the two tasks.}
    \label{fig:objective-main}
\end{figure}
\addtocounter{figure}{1}%

\subsection{Experimental Setup}
\label{sec:units:exp_setup}

\paragraph{Evaluation Metrics.} We use \textbf{accuracy} for both the background inference and question answering tasks. Since our task is generative, we use a regex search to match the nationality (e.g., ``France'') and the desired unit (e.g., °C); see Appendix~\ref{app: gt units}.

\paragraph{Models.} To manage cost, we evaluate a single closed-source model, Gemini-3.1-Flash-Lite. This model fits our purpose: it is designed for fast responses in daily conversation, while also supporting chain-of-thought reasoning \cite{wei2022chain}. 
We evaluate a broader set of open-source models, spanning different organizations and countries of origin: Qwen3-VL-8B and 32B, in both -Instruct and -Thinking variants \cite{bai2025qwen3}; Llama-3.2-11B-Vision-Instruct (direct prediction only) \cite{grattafiori2024llama}; and Gemma-4 in sizes E2B, E4B, and 31B, all of which support both direct and thinking modes \cite{team2026gemma} (details in Appendix~\ref{app: model details}). 

\paragraph{Implementation Details.} We employ a minimal prompt: (1) for the primary VQA task, we ask the models to role-play the chatbot and answer the user's question in line with their cultural background; (2) for the background inference task, we simply ask the models to infer the background (prompts detailed in Appendix~\ref{app: prompt details}). We set \texttt{temperature=0} across all tasks for reproducibility. All models are hosted on vLLM \cite{kwon2023efficient} to accelerate inference, except LLaMA, which we host via HuggingFace \cite{wolf-etal-2020-transformers} for compatibility reasons. For reasoning models, we set a maximum length of 2048 tokens, which is sufficient for our task, detailed in Appendix~\ref{app: inference details}.

\subsection{Overall Performance (RQ1)}
\label{sec:units:overall}

We present the best model per family as a function of number of cues in Figure~\ref{fig:objective-main}. Each point aggregates $n=3{,}672$ conversations across the ten countries; 95\% CIs are about $\pm1.6$ points, smaller than the marker size. At the start, background inference (BG, solid line) is near zero since there are no cultural cues. The VQA task (unit prediction, dashed line) sits around 30-60\%, since models can guess a plausible answer from the image.
A clear trend emerges: with just one cultural cue (from \neutral to \implicitcueone), background prediction jumps sharply from near zero to around 80\%, while the VQA task rises much more slowly. The gap widens with more cues (\implicitcuetwo and \implicitcuefull): background prediction is almost perfect, while the VQA task still lags behind. \textbf{This suggests current LLMs are not ideal pragmatic speakers:} they have the capacity to infer the user's background correctly, but do not use it when answering. Finally, we use \explicit as an upper bound: with the background stated explicitly, background prediction is perfect and the gap narrows for all models except LLaMA, which appears unable to associate the cultural background with the correct units.

\begin{figure*}[!tbp]
    \centering
    \includegraphics[width=0.9\textwidth]{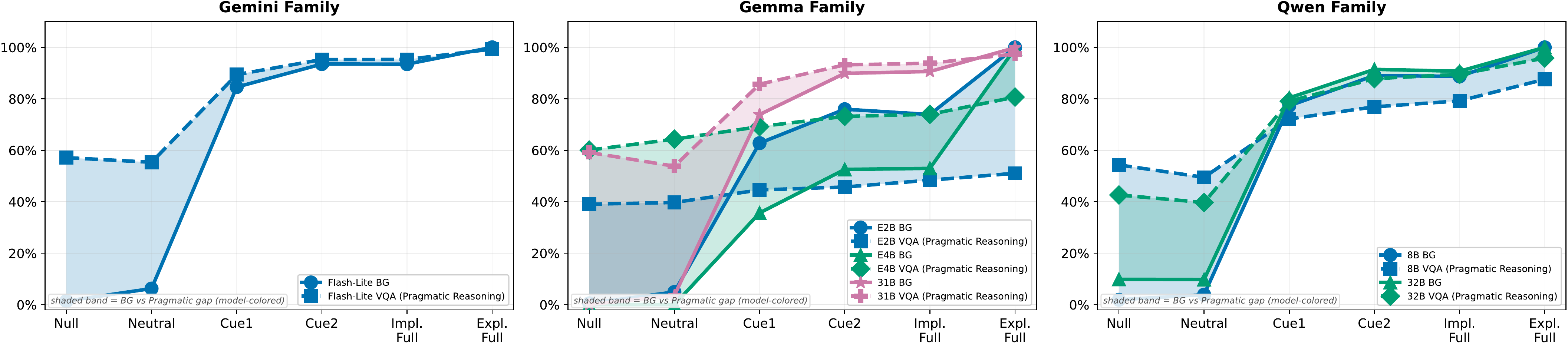}
    \caption{\textbf{RQ3:} Effect of model scaling across the Qwen, Gemma, and Gemini-3.1 families: Larger models narrow the gap between VQA and BG (the shaded area).}
    \label{fig:result-1-3}
\end{figure*}

\subsection{Benefit of Reasoning (RQ2)}
\label{sec:units:reasoning}

We then show that CoT reasoning significantly improves VQA performance, with Pragmatic CoT bringing the largest gain (best reasoning model per family, reported in Figure~\ref{fig:result-1-2}).
\textit{Ideally}, the model should (1) infer the user's background, then (2) produce an answer aligned with that background. We test whether models can perform such pragmatic reasoning via CoT, exploring two zero-shot setups: 
(1) \textbf{CoT}, the plain version where the model uses its default reasoning mode with no culture-specific guidance (it must reason from the task instruction alone);
(2) \textbf{Pragmatic CoT}, where we add explicit instructions for the model to infer the cultural background before making a prediction.
Figure~\ref{fig:result-1-2} shows that plain CoT already improves over direct VQA, while Pragmatic CoT improves VQA performance substantially, matching or exceeding the background inference task. \textbf{Together, these results show that with explicit reasoning, models can perform pragmatic reasoning in cultural contexts.}

Further analyses show interesting behaviors of models' reasoning: (1) Pragmatic CoT produces deeper reasoning, often linking the inferred background to the answer via an explicit causal connective. (2) Plain CoT yields shallower reasoning overall, though larger models still reach the deeper levels, suggesting that pragmatic reasoning capability scales with model size
(detailed in Appendix~\ref{app: reasoning depth}).

\subsection{Benefit of Scaling (RQ3)}
\label{sec:units:size}
We further find that larger models close the gap between BG and VQA more effectively under Pragmatic CoT (Figure~\ref{fig:result-1-3}).
(1) Starting with the strongest overall model, Gemini-3.1, we see that with just one cue (\implicitcueone), the gap between VQA and BG is already very small, and it diminishes further by the end. This shows that with proper Pragmatic CoT guidance, Gemini-3.1 handles the reasoning well. (2) Turning to the Qwen and Gemma families, we see a clear trend: larger models have a smaller gap between VQA and BG. The shaded shape of Qwen-32B and Gemma-31B resembles Gemini-3.1; however, the gap is significantly larger for the smaller Qwen and Gemma models, suggesting that \textbf{LLMs have an intrinsic limitation in pragmatic reasoning in cultural contexts that even explicit guidance cannot overcome}.

\subsection{Models' Cultural Prior (RQ4)}
\label{sec:units:prior}

\begin{figure}[!ht]
    \centering
    \begin{subfigure}{0.95\columnwidth}
        \centering
        \includegraphics[width=\linewidth]{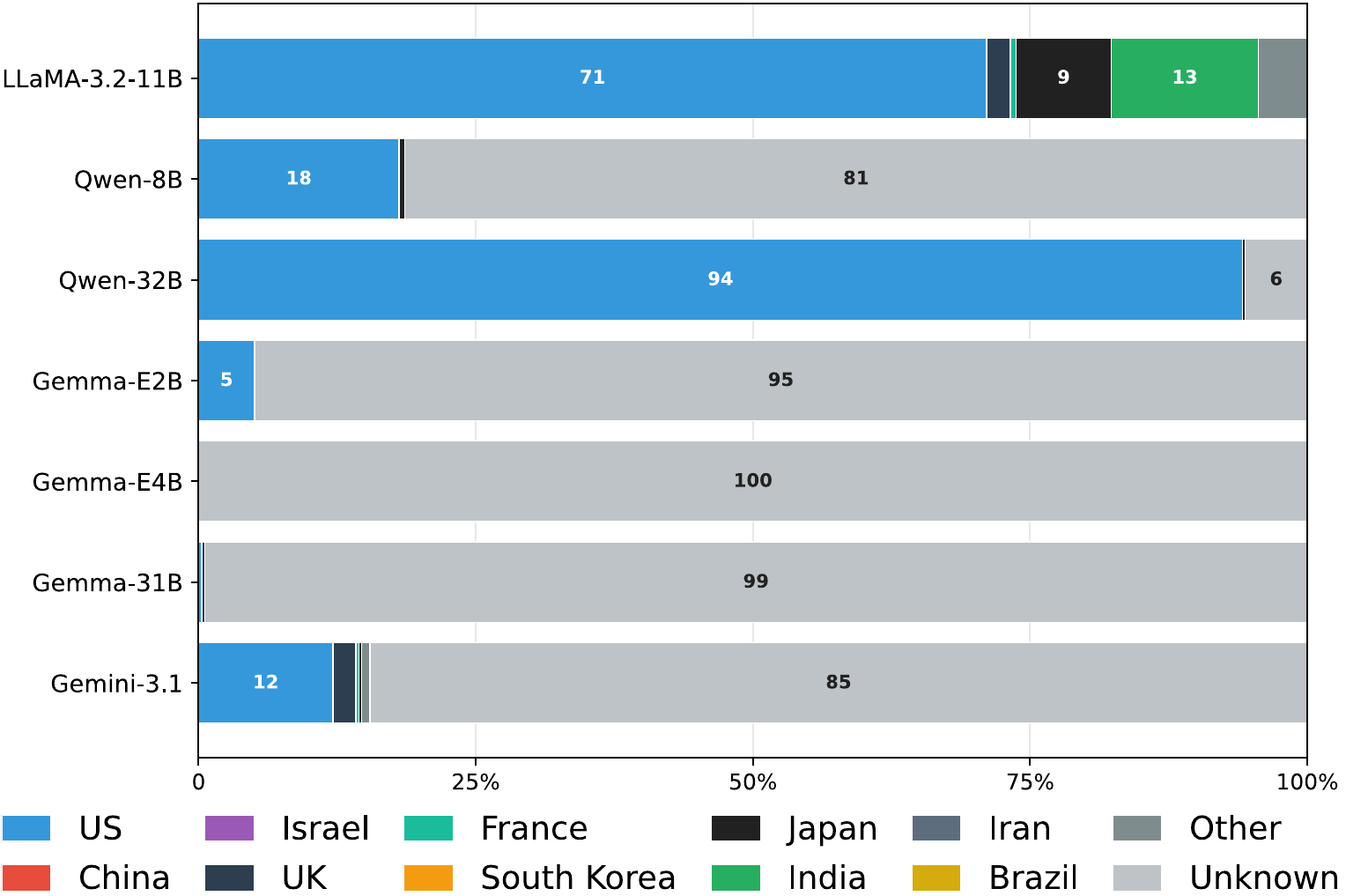}
        \caption{Background prior (BG): models lean toward the US.}
        \label{fig:prior-id-stacked}
    \end{subfigure}

    \vspace{0.5em}

    \begin{subfigure}{0.95\columnwidth}
        \centering
        \includegraphics[width=\linewidth]{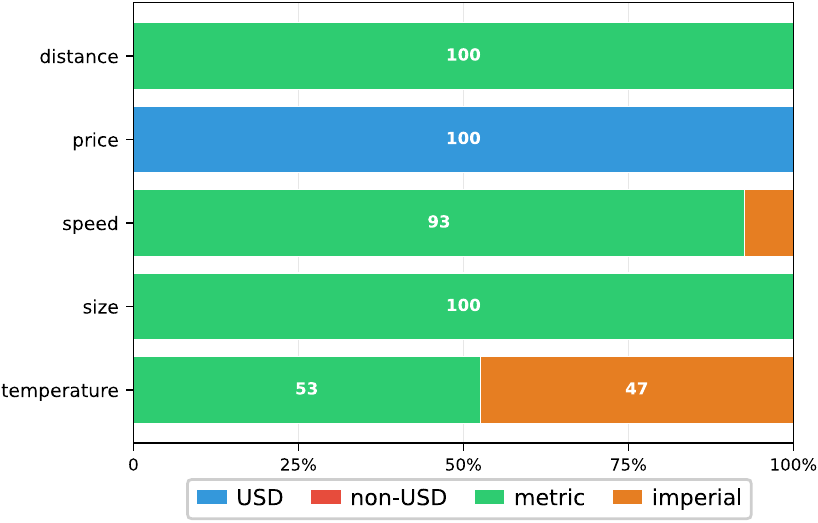}
        \caption{Unit prior (VQA): models lean toward the metric system.}
        \label{fig:prior-vqa-gemini}
    \end{subfigure}
    \caption{\textbf{RQ4:} Models' cultural prior at \nullcue, across the two tasks.}
    \label{fig:prior-rq4}
\end{figure}

\textbf{Finally, we examine the models' cultural priors: without any cues, models default to the US for country, yet prefer the metric system (which the US does not use) for measurement units.}
Users might ask models a question without any contextual cues, so it is important to quantify models' cultural prior. We treat models' responses on \nullcue conversations as the prior. Figure~\ref{fig:prior-id-stacked} shows all models' prior on the background prediction task. From a bird's-eye view, the majority response is \texttt{unknown} (the model explicitly refuses to answer). The most popular predicted country is the US (by LLaMA, Qwen, and Gemini-3.1). Interestingly, Qwen-8B refuses to answer, while Qwen-32B becomes overconfident in predicting the US (94\%). LLaMA and Gemini-3.1 are the most ``diverse'' models, with notable shares for Japan, UK, India, and others.

Models' prior on unit prediction tells a different story (Figure~\ref{fig:prior-vqa-gemini} shows Gemini-3.1; other models are similar). Although background prediction leans toward the US, unit predictions are dominated by the metric system, the opposite of US convention. While the metric system is far more common around the world than the imperial system, this relationship is not straightforwardly captured in term frequency in large web corpora. A search of infini-gram \cite{liuinfini} showed that imperial and metric size units appear roughly equally, whereas distance and temperature have preference for metric units ($\sim$60\%) and speed has a more substantial preference for imperial ($\sim$70\%).  

Models prediction for pricing is an exception, where the dollar dominates (100\% for Gemini, other models similar). Currencies differ across all countries in our data, and USD is more frequently discussed in English text than other currencies. A search of infini-gram resulted in 113M hits for ``USD'', double the number of hits as for all other currencies tested in this paper, combined (58M). 

\section{Do LLMs condition on culture when interpreting subjective expressions?}
\label{sec:subjective}
To what extent are LLM outputs on subjective dimensions -- such as mapping 6 PM to ``evening'' or ``afternoon'' --  sensitive to cultural context, rather than reflecting a single dominant or default interpretation?
Using the same models as in Sec.~\ref{sec:units:exp_setup}, we address the following research questions:

\begin{tightitemize}
\item \textbf{RQ5:} How do cultural cues influence models' subjective predictions (\S\ref{sec:subjective:cue_influence})?
\item \textbf{RQ6:} Which cultures do models' priors lean toward (\S\ref{sec:subjective:specific})?
\end{tightitemize}

\subsection{Cultural Cues' Influence (RQ5)}
\label{sec:subjective:cue_influence}

\begin{figure}[h]
    \centering
    \includegraphics[width=0.95\columnwidth]{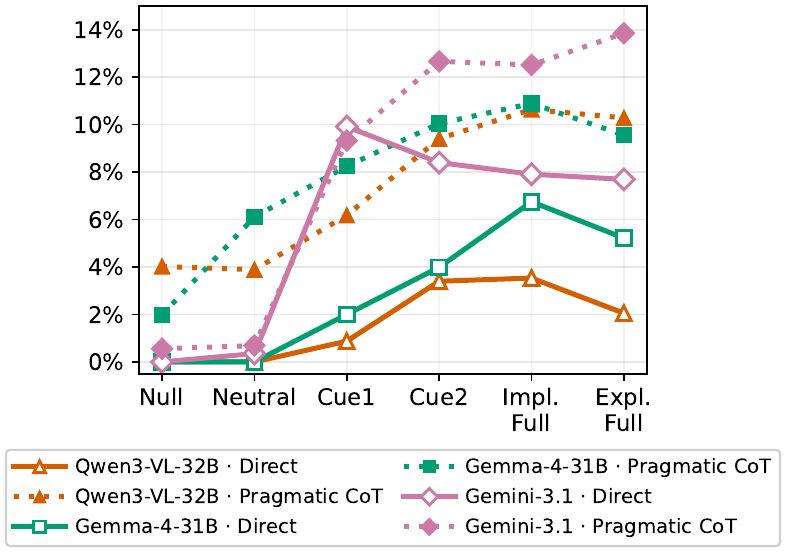}
    \caption{\textbf{RQ5:} The impact of cultural cues on models' inter-culture divergence ($D_{\mathrm{Inter}}$): more cues yield more diverse answers.}
    \label{fig:result-2-main}
\end{figure}

We first study whether models show more divergence as cues accumulate. Given that there is no ground-truth answer for subjective dimensions, we instead examine how models' predictions shift as cultural cues accumulate, by defining the \textbf{inter-culture divergence score}:
$$ D_{\mathrm{Inter}} = \frac{1}{\binom{N}{2}} \sum_{\{B,B'\}} \Pr(y^{B} \neq y^{B'}) $$
\noindent $D_{\mathrm{Inter}}$ is the average divergence across all $\binom{N}{2}=45$ pairs of cultures (with $N=10$) at the same cue level, where $B$ and $B'$ denote two distinct cultural backgrounds. 
For example, if a model interpreted 6 PM as ``afternoon'' under one cultural condition and as ``evening'' under another, we count this as a divergence. We hold the conversation fixed and vary only the cultural cues across countries: a [shopping platform] cue might be \textit{Fnac} (France), \textit{Flipkart} (India), or \textit{Coupang} (South Korea). If a model is sensitive to these cues, it may give different answers to reflect the common interpretation within a given culture. We use this divergence as a measure of cultural specificity (visualized in Appendix~\ref{app: subjective vis}).

Fig.~\ref{fig:result-2-main} reports each family's largest model on the subjective task, measured by $D_{\mathrm{Inter}}$. We make two observations: (1) Overall, models give more diverse answers as cues accumulate, from \implicitcueone to \implicitcuetwo to \implicitcuefull. Interestingly, for some models the divergence drops from \implicitcuefull to \explicit.
(2) Pragmatic CoT increases divergence compared to direct prediction. Inspecting the reasoning traces, we find that stronger models follow our instructions and engage in deliberate thinking such as ``the user is from \dots, so what does 6 PM mean to them?'', which enhances the divergence.

\definecolor{ColorDown}{HTML}{2166AC}
\definecolor{ColorUp}{HTML}{D6604D}

\begin{table}[t]
  \centering
  \small
  \setlength{\tabcolsep}{4pt}
  \begin{tabular*}{\columnwidth}{@{\extracolsep{\fill}} l rr rr rr @{}}
  \toprule
   & \multicolumn{2}{c}{\textbf{Qwen-32B}} & \multicolumn{2}{c}{\textbf{Gemma-31B}} & \multicolumn{2}{c}{\textbf{Gemini-3.1}} \\
  \cmidrule(lr){2-3}\cmidrule(lr){4-5}\cmidrule(lr){6-7}
  \textbf{Culture} & \textbf{$D_{\mathrm{Impl}}$} & \textbf{$D_{\mathrm{Expl}}$} & \textbf{$D_{\mathrm{Impl}}$} & \textbf{$D_{\mathrm{Expl}}$} & \textbf{$D_{\mathrm{Impl}}$} & \textbf{$D_{\mathrm{Expl}}$} \\
  \midrule
  US & 6.8 & \textcolor{ColorUp}{3.8\,$\uparrow$} & 8.3 & \textcolor{ColorUp}{9.8\,$\uparrow$} & 9.8 & 6.8 \\
  China & \textcolor{ColorDown}{4.5\,$\downarrow$} & \textcolor{ColorDown}{1.5\,$\downarrow$} & \textcolor{ColorDown}{7.6\,$\downarrow$} & 7.6 & 8.3 & \textcolor{ColorUp}{11.4\,$\uparrow$} \\
  Israel & 5.3 & \textcolor{ColorDown}{1.5\,$\downarrow$} & 9.1 & 8.3 & 6.1 & 8.3 \\
  UK & 5.3 & 3.0 & 8.3 & 6.1 & 7.6 & 9.1 \\
  France & \textcolor{ColorUp}{7.6\,$\uparrow$} & 3.0 & \textcolor{ColorUp}{9.8\,$\uparrow$} & 7.6 & 8.3 & 9.8 \\
  Korea & 5.3 & 3.0 & 9.1 & 6.1 & \textcolor{ColorDown}{5.3\,$\downarrow$} & 10.6 \\
  Japan & 6.1 & 2.3 & \textcolor{ColorDown}{7.6\,$\downarrow$} & 6.8 & 7.6 & 8.3 \\
  India & \textcolor{ColorDown}{4.5\,$\downarrow$} & 2.3 & \textcolor{ColorDown}{7.6\,$\downarrow$} & \textcolor{ColorDown}{5.3\,$\downarrow$} & 9.1 & 8.3 \\
  Iran & \textcolor{ColorDown}{4.5\,$\downarrow$} & 3.0 & 8.3 & 9.1 & 6.1 & 7.6 \\
  Brazil & 6.1 & 2.3 & 9.1 & 6.8 & \textcolor{ColorUp}{10.6\,$\uparrow$} & \textcolor{ColorDown}{6.1\,$\downarrow$} \\
  \cmidrule(lr){2-7}
  \textit{Mean} & 5.6 & 2.6 & 8.5 & 7.3 & 7.9 & 8.6 \\
  \bottomrule
  \end{tabular*}
  \caption{\textbf{RQ6:} Per-culture $D_{\mathrm{Impl}}$ and $D_{\mathrm{Expl}}$ for three models (in percentage score). \textcolor{ColorDown}{$\downarrow$} marks the per-column minimum (culture closest to the model's prior); \textcolor{ColorUp}{$\uparrow$} marks the per-column maximum (the most ``alien'' culture). 95\% CIs are about $\pm1.5$ for the \textit{Mean} row ($n=1{,}320$) and $\pm4.3$ for individual cultures ($n=132$).}
  \label{tab:variation-score}
\end{table}

\subsection{Sensitivity to Specific Cultures (RQ6)}
\label{sec:subjective:specific}

We then test models' cultural priors on these subjective concepts, where the absence of a ground-truth answer lets each model's default leaning emerge. To that end, we define the \textbf{cue divergence scores}:
$$ D_{\mathrm{Expl}}(B) = \Pr(y^{B}_{\textsc{ExplicitFull}} \neq y^{B}_{\textsc{Neutral}}) $$
$$ D_{\mathrm{Impl}}(B) = \Pr(y^{B}_{\textsc{ImplicitFull}} \neq y^{B}_{\textsc{Neutral}}) $$
\noindent where $D_{\mathrm{Expl}}(B)$ and $D_{\mathrm{Impl}}(B)$ are the \textbf{explicit-} and \textbf{implicit-cue divergence scores} for culture $B$. To attribute a shift cleanly to the cultural cues, we anchor on the \neutral conversation and construct two minimally-changed variants: \explicit only appends an ``I am from \dots'' phrase to \neutral, while \implicitcuefull only fills the neutralized \texttt{[cue]} slots in \neutral with culture-specific values. Each score measures, per culture, how often the model's answer changes under that minimal modification. Low scores may indicate that the model's priors align with the given culture. 

We report the results in Table~\ref{tab:variation-score}. The per-culture cells carry wide confidence intervals, so we see them as trends rather than definitive rankings. Three findings stand out: (1) The US is the most \textcolor{ColorUp}{alien} culture for Qwen-32B and for Gemma-31B ($D_{\mathrm{Expl}}$), and is never the \textcolor{ColorDown}{closest} to any model's prior. (2) China is the \textcolor{ColorDown}{closest} to Qwen-32B's prior, consistent with Qwen being developed by a Chinese company, and is also favored by Gemma-31B under implicit cues. Interestingly, Gemini-3.1 instead places China as its most \textcolor{ColorUp}{alien} culture under explicit cues. (3) For Gemini-3.1, Brazil is the \textcolor{ColorDown}{closest} under explicit cues but the most \textcolor{ColorUp}{alien} under implicit cues. Possible reason is that Brazil shares implicit cues with other Latin American countries, which can confuse the model (case study in Appendix~\ref{app: case study}).

\section{Related Work}
\label{sec:related_work}
\paragraph{Pragmatic Speaker Models.} A pragmatic speaker aims to maximize communicative effect by inferring how the listener interprets each utterance \cite{frank2012predicting, andreas-klein-2016-reasoning}. This process often involves resolving implicatures \cite{ruis2023the, cong2024manner}.
Recent studies show that LLMs still fall short at pragmatic understanding \cite{hu-etal-2023-fine, sravanthi-etal-2024-pub}, and a significant gap remains before they can reason as competent pragmatic speakers \cite{jian2024llmsgoodpragmaticspeakers, sieker-zarriess-2026-hypocritical}.
Pragmatic reasoning has been applied to many tasks, such as code generation \cite{cao2025pragmatic}, multi-turn dialogue \cite{estienne-etal-2025-collaborative}, and image captioning \cite{cohn-gordon-etal-2018-pragmatically, nie-etal-2020-pragmatic}, but cultural aspects, our main focus, are largely overlooked in this line of work. 
A notable exception is \citet{white-etal-2024-communicate}, who study cross-cultural common ground between two players in Codenames Duet, a cooperative word-association game. Our setting differs in both task and direction: we focus on concept-to-value grounding (e.g., units, time expressions) and on user-to-model adaptation, where the user's culture is inferred from cues.

\paragraph{Cultural Competence in LLMs and VLMs.} LLMs are expected to respond in line with the speaker's cultural background. Recent work largely focuses on evaluating LLMs' cultural competence with respect to cultural values \cite{zhao-etal-2024-worldvaluesbench, kabir-etal-2025-break}, norms \cite{rao-etal-2025-normad}, and knowledge \cite{ramezani-xu-2023-knowledge, myung2024blend, chiu-etal-2025-culturalbench, mor-lan-etal-2026-location, naous-etal-2024-beer,veselovsky-etal-2026-localized}.
These evaluations have been extended to VLMs and text-to-image models \cite{bhatia-etal-2024-local,nayak-etal-2024-benchmarking,nayak-etal-2025-culturalframes}.
Many recent resources target cross-cultural understanding, including multimodal metaphor \cite{yang-etal-2025-cultural}, emotion understanding \cite{belay-etal-2025-culemo}, cross-regional object recognition \cite{rojas2022the}, as well as Hofstede's cultural dimensions \cite{kharchenko2024well}. However, they overlook how the expression of the same concept varies across cultures.

\paragraph{Language Grounding}

Grounding is the process of mapping utterances to what they refer to in the world, for example, the expression ``evening'' to a specific time, or ``few'' to a particular count (e.g., of eggs in a basket). We take a novel angle: grounding in images, leaving the culturally appropriate lexical choice to the model. Language grounding has been studied across text, image, and video \cite{chandu-etal-2021-grounding, fried-etal-2023-pragmatics}, but less so in cultural contexts. Notable exceptions include time expressions \cite{shwartz-2022-good}, quantifiers \cite{stateva2019cross, wong-etal-2025-vaquum}, gradable adjectives \cite{gari-soler-apidianaki-2021-scalar}, and implicit numeric heads \cite{elazar-goldberg-2019-wheres}. However, these resources either treat concepts as culture-invariant or capture cross-cultural variation only through small-scale psycholinguistic studies. Therefore we introduce \datasetname, which grounds both objective and subjective concepts across cultures at scale. Our evaluation is also distinct in a conversational setting, disentangling what models know from whether they act on it.

\section{Conclusion}
\label{sec:conclusion}
Cultural competence in LLMs is becoming increasingly consequential as they are deployed to automate processes globally. 
We introduce \datasetname{},  
reframing the question from whether LLMs \textit{know} a certain culture to whether they can \textit{apply} this knowledge and tailor their answer to maximize communicative effect.  
Our evaluation reveals a persistent gap: state-of-the-art models reliably infer the user's background but fail to apply it, unless explicitly guided to reason about this. 
We also show that on subjective expressions, models' answers diverge more as cultural cues accumulate, while their no-cue priors sometimes align with the model's country of origin.
\datasetname{} is a step toward narrowing the gap between cultural knowledge and culturally adaptive generation. Future work should broaden the set of grounding dimensions and cultures, and imbue this reasoning capability into LLMs, making them pragmatic speakers.

\section*{Limitations}
\label{sec:limitations}
\paragraph{Coverage of cultures.} We follow the common practice in the NLP community and use country as a proxy for culture \cite{wang-etal-2024-countries,liu-etal-2025-culturally}, while acknowledging that cultures could be defined at a finer-grained level based on region, language, religion, and more. While we only tested 10 countries, we selected them to maximize regional diversity. 

\paragraph{Language artifacts.} Our dataset is in English, which introduces a language prior. We chose English to separate the effect of LLM performance on different languages from their ability to perform cultural pragmatic reasoning. 
A small case study on translated conversations (Appendix~\ref{app: translated conversation}) shows that the cultural reasoning is still not perfect: the conversation's language alone makes background inference nearly trivial, while the gap to question answering persists.
We leave full multilingual evaluation, where the input language itself can signal the user's background, to future work. 

\section*{Ethical Considerations}
\label{sec:ethics}
\paragraph{Annotation.} The study was conducted with approval from our institute's Behavioral Research Ethics Board (IRB). Annotation is performed on the Cloud Connect platform, where we pay annotators \$15 USD/hour, in line with CloudResearch's compensation guidelines to pay at least the local minimum wage standards. 

\paragraph{Data Sources and Potential Risks.} We generated the conversations with Google Gemini-2.5-Pro. We generated the images with Google Imagen 4 and Gemini-2.5-Flash-Image, apart from photographs, which we sourced from Flickr under permissive licenses (``commercial use \& modifications allowed''). The cues (names, phone numbers, postal codes) are synthetic and follow each country's format conventions; they do not refer to any real individual. The images are about objects and scenes (thermometers, rooms, tractors, etc.) rather than identifiable people, and we manually inspect each to filter out any with explicit text or strong cultural signals. The conversation topics cover neutral concepts (measurement units, time, quantifiers), avoiding sensitive or harmful content.

\paragraph{Cultural Biases.} In this work, we use country-level background as a proxy for culture, following common practice in NLP research. We mitigate stereotyping by verifying both the cultural cues and the conversations with people from the target country (our annotators have lived there for at least 5 of the past 15 years; \S\ref{sec:dataset:creation}). However, this simplification may still reinforce the stereotype that people from the same country share a homogeneous culture. Future work could consider more fine-grained representations of culture beyond country-level proxies.

\paragraph{AI Tool Usage.} We use Google Gemini-2.5-Pro for conversation generation. For image generation, we use Google Imagen 4 and Gemini-2.5-Flash-Image (\S\ref{sec:dataset:creation}). Claude Code and Cursor AI were used during coding, primarily for debugging assistance; ChatGPT and Claude (via the web interface) was used only for grammar checking of the manuscript.

\section*{Acknowledgements}
\label{sec:ack}
We thank anonymous reviewers for spending time in reviewing our paper, and for their helpful feedback. 
We thank Aditya Chinchure, Soheil Alavi, Eunjeong Hwang, Sara Papi, Samuel Rhys Cox, and Hannah Brown for help fact-checking our knowledge base, and Joy Zhuozhuo Liu for her annotation interface that inspired our design. 
Yisong thanks Dr. Jim Mondo for his IMPACT Program mentorship during his Vector internship. 
We also thank Prof. Min-Yen Kan for following our updates and offering insightful comments. 

This work was supported by the Vector Institute and was conducted during the first author’s internship at the institute. It was also supported by funding from UBC Language Sciences. This research was enabled in part by compute credits from Google for Gemini. The experiments were partially supported by NUS HPC clusters. The authors are also supported by grants from NSERC and the Canada CIFAR AI Chairs program.


\bibliography{anthology-1,anthology-2,custom}

\begin{thebibliography}{55}
\providecommand{\natexlab}[1]{#1}

\bibitem[{Andreas and Klein(2016)}]{andreas-klein-2016-reasoning}
Jacob Andreas and Dan Klein. 2016.
\newblock \href {https://doi.org/10.18653/v1/D16-1125} {Reasoning about pragmatics with neural listeners and speakers}.
\newblock In \emph{Proceedings of the 2016 Conference on Empirical Methods in Natural Language Processing}, pages 1173--1182, Austin, Texas. Association for Computational Linguistics.

\bibitem[{Bai et~al.(2025)Bai, Cai, Chen, Chen, Chen, Cheng, Deng, Ding, Gao, Ge, Ge, Guo, Huang, Huang, Huang, Hui, Jiang, Li, Li, Li, Li, Lin, Lin, Liu, Liu, Liu, Liu, Liu, Liu, Lu, Luo, Lv, Men, Meng, Ren, Ren, Song, Sun, Tang, Tu, Wan, Wang, Wang, Wang, Wang, Xie, Xu, Xu, Xu, Yang, Yang, Yang, Yang, Yu, Zhang, Zhang, Zhang, Zheng, Zhong, Zhou, Zhou, Zhou, Zhu, and Zhu}]{bai2025qwen3}
Shuai Bai, Yuxuan Cai, Ruizhe Chen, Keqin Chen, Xionghui Chen, Zesen Cheng, Lianghao Deng, Wei Ding, Chang Gao, Chunjiang Ge, Wenbin Ge, Zhifang Guo, Qidong Huang, Jie Huang, Fei Huang, Binyuan Hui, Shutong Jiang, Zhaohai Li, Mingsheng Li, and 45 others. 2025.
\newblock \href {https://arxiv.org/abs/2511.21631} {Qwen3-vl technical report}.
\newblock \emph{Preprint}, arXiv:2511.21631.

\bibitem[{Belay et~al.(2025)Belay, Ahmed, Grissom~II, Ameer, Sidorov, Kolesnikova, and Yimam}]{belay-etal-2025-culemo}
Tadesse~Destaw Belay, Ahmed~Haj Ahmed, Alvin Grissom~II, Iqra Ameer, Grigori Sidorov, Olga Kolesnikova, and Seid~Muhie Yimam. 2025.
\newblock \href {https://doi.org/10.18653/v1/2025.acl-long.925} {{CULEMO}: Cultural lenses on emotion - benchmarking {LLM}s for cross-cultural emotion understanding}.
\newblock In \emph{Proceedings of the 63rd Annual Meeting of the Association for Computational Linguistics (Volume 1: Long Papers)}, pages 18894--18909, Vienna, Austria. Association for Computational Linguistics.

\bibitem[{Bhatia et~al.(2024)Bhatia, Ravi, Chinchure, Hwang, and Shwartz}]{bhatia-etal-2024-local}
Mehar Bhatia, Sahithya Ravi, Aditya Chinchure, EunJeong Hwang, and Vered Shwartz. 2024.
\newblock \href {https://doi.org/10.18653/v1/2024.emnlp-main.385} {From local concepts to universals: Evaluating the multicultural understanding of vision-language models}.
\newblock In \emph{Proceedings of the 2024 Conference on Empirical Methods in Natural Language Processing}, pages 6763--6782, Miami, Florida, USA. Association for Computational Linguistics.

\bibitem[{Cao et~al.(2024)Cao, Chen, and Hershcovich}]{cao-etal-2024-bridging}
Yong Cao, Min Chen, and Daniel Hershcovich. 2024.
\newblock \href {https://doi.org/10.18653/v1/2024.findings-eacl.63} {Bridging cultural nuances in dialogue agents through cultural value surveys}.
\newblock In \emph{Findings of the Association for Computational Linguistics: EACL 2024}, pages 929--945, St. Julian{'}s, Malta. Association for Computational Linguistics.

\bibitem[{Cao et~al.(2025{\natexlab{a}})Cao, Liu, Arora, Augenstein, R{\"o}ttger, and Hershcovich}]{cao-etal-2025-specializing}
Yong Cao, Haijiang Liu, Arnav Arora, Isabelle Augenstein, Paul R{\"o}ttger, and Daniel Hershcovich. 2025{\natexlab{a}}.
\newblock \href {https://doi.org/10.18653/v1/2025.naacl-long.162} {Specializing large language models to simulate survey response distributions for global populations}.
\newblock In \emph{Proceedings of the 2025 Conference of the Nations of the Americas Chapter of the Association for Computational Linguistics: Human Language Technologies (Volume 1: Long Papers)}, pages 3141--3154, Albuquerque, New Mexico. Association for Computational Linguistics.

\bibitem[{Cao et~al.(2023)Cao, Zhou, Lee, Cabello, Chen, and Hershcovich}]{cao-etal-2023-assessing}
Yong Cao, Li~Zhou, Seolhwa Lee, Laura Cabello, Min Chen, and Daniel Hershcovich. 2023.
\newblock \href {https://doi.org/10.18653/v1/2023.c3nlp-1.7} {Assessing cross-cultural alignment between {C}hat{GPT} and human societies: An empirical study}.
\newblock In \emph{Proceedings of the First Workshop on Cross-Cultural Considerations in NLP (C3NLP)}, pages 53--67, Dubrovnik, Croatia. Association for Computational Linguistics.

\bibitem[{Cao et~al.(2025{\natexlab{b}})Cao, Apel, Singla, and Demberg}]{cao2025pragmatic}
Zhuchen Cao, Sven Apel, Adish Singla, and Vera Demberg. 2025{\natexlab{b}}.
\newblock Pragmatic reasoning improves llm code generation.
\newblock \emph{arXiv preprint arXiv:2502.15835}.

\bibitem[{Chandu et~al.(2021)Chandu, Bisk, and Black}]{chandu-etal-2021-grounding}
Khyathi~Raghavi Chandu, Yonatan Bisk, and Alan~W Black. 2021.
\newblock \href {https://doi.org/10.18653/v1/2021.findings-acl.375} {Grounding `grounding' in {NLP}}.
\newblock In \emph{Findings of the Association for Computational Linguistics: ACL-IJCNLP 2021}, pages 4283--4305, Online. Association for Computational Linguistics.

\bibitem[{Chiu et~al.(2025)Chiu, Jiang, Lin, Park, Li, Ravi, Bhatia, Antoniak, Tsvetkov, Shwartz, and Choi}]{chiu-etal-2025-culturalbench}
Yu~Ying Chiu, Liwei Jiang, Bill~Yuchen Lin, Chan~Young Park, Shuyue~Stella Li, Sahithya Ravi, Mehar Bhatia, Maria Antoniak, Yulia Tsvetkov, Vered Shwartz, and Yejin Choi. 2025.
\newblock \href {https://doi.org/10.18653/v1/2025.acl-long.1247} {{C}ultural{B}ench: A robust, diverse and challenging benchmark for measuring {LM}s' cultural knowledge through human-{AI} red-teaming}.
\newblock In \emph{Proceedings of the 63rd Annual Meeting of the Association for Computational Linguistics (Volume 1: Long Papers)}, pages 25663--25701, Vienna, Austria. Association for Computational Linguistics.

\bibitem[{Cohn-Gordon et~al.(2018)Cohn-Gordon, Goodman, and Potts}]{cohn-gordon-etal-2018-pragmatically}
Reuben Cohn-Gordon, Noah Goodman, and Christopher Potts. 2018.
\newblock \href {https://doi.org/10.18653/v1/N18-2070} {Pragmatically informative image captioning with character-level inference}.
\newblock In \emph{Proceedings of the 2018 Conference of the North {A}merican Chapter of the Association for Computational Linguistics: Human Language Technologies, Volume 2 (Short Papers)}, pages 439--443, New Orleans, Louisiana. Association for Computational Linguistics.

\bibitem[{Cong(2024)}]{cong2024manner}
Yan Cong. 2024.
\newblock Manner implicatures in large language models.
\newblock \emph{Scientific Reports}, 14(1):29113.

\bibitem[{DURMUS et~al.(2024)DURMUS, Nguyen, Liao, Schiefer, Askell, Bakhtin, Chen, Hatfield-Dodds, Hernandez, Joseph, Lovitt, McCandlish, Sikder, Tamkin, Thamkul, Kaplan, Clark, and Ganguli}]{durmustowards}
Esin DURMUS, Karina Nguyen, Thomas Liao, Nicholas Schiefer, Amanda Askell, Anton Bakhtin, Carol Chen, Zac Hatfield-Dodds, Danny Hernandez, Nicholas Joseph, Liane Lovitt, Sam McCandlish, Orowa Sikder, Alex Tamkin, Janel Thamkul, Jared Kaplan, Jack Clark, and Deep Ganguli. 2024.
\newblock \href {https://openreview.net/forum?id=zl16jLb91v} {Towards measuring the representation of subjective global opinions in language models}.
\newblock In \emph{First Conference on Language Modeling}.

\bibitem[{Elazar and Goldberg(2019)}]{elazar-goldberg-2019-wheres}
Yanai Elazar and Yoav Goldberg. 2019.
\newblock \href {https://doi.org/10.1162/tacl_a_00280} {Where{'}s my head? {D}efinition, data set, and models for numeric fused-head identification and resolution}.
\newblock \emph{Transactions of the Association for Computational Linguistics}, 7:519--535.

\bibitem[{Estienne et~al.(2025)Estienne, Zenou, Naderi, Cheung, and Piantanida}]{estienne-etal-2025-collaborative}
Lautaro Estienne, Gabriel~Ben Zenou, Nona Naderi, Jackie~CK Cheung, and Pablo Piantanida. 2025.
\newblock \href {https://doi.org/10.18653/v1/2025.emnlp-main.1145} {Collaborative rational speech act: Pragmatic reasoning for multi-turn dialog}.
\newblock In \emph{Proceedings of the 2025 Conference on Empirical Methods in Natural Language Processing}, pages 22509--22523, Suzhou, China. Association for Computational Linguistics.

\bibitem[{Frank and Goodman(2012)}]{frank2012predicting}
Michael~C Frank and Noah~D Goodman. 2012.
\newblock Predicting pragmatic reasoning in language games.
\newblock \emph{Science}, 336(6084):998--998.

\bibitem[{Fried et~al.(2023)Fried, Tomlin, Hu, Patel, and Nematzadeh}]{fried-etal-2023-pragmatics}
Daniel Fried, Nicholas Tomlin, Jennifer Hu, Roma Patel, and Aida Nematzadeh. 2023.
\newblock \href {https://doi.org/10.18653/v1/2023.findings-emnlp.840} {Pragmatics in language grounding: Phenomena, tasks, and modeling approaches}.
\newblock In \emph{Findings of the Association for Computational Linguistics: EMNLP 2023}, pages 12619--12640, Singapore. Association for Computational Linguistics.

\bibitem[{Gar{\'i}~Soler and Apidianaki(2021)}]{gari-soler-apidianaki-2021-scalar}
Aina Gar{\'i}~Soler and Marianna Apidianaki. 2021.
\newblock \href {https://doi.org/10.18653/v1/2021.naacl-main.370} {Scalar adjective identification and multilingual ranking}.
\newblock In \emph{Proceedings of the 2021 Conference of the North American Chapter of the Association for Computational Linguistics: Human Language Technologies}, pages 4653--4660, Online. Association for Computational Linguistics.

\bibitem[{Grattafiori et~al.(2024)Grattafiori, Dubey, Jauhri, Pandey, Kadian, Al-Dahle, Letman, Mathur, Schelten, Vaughan, Yang, Fan, Goyal, Hartshorn, Yang, Mitra, Sravankumar, Korenev, Hinsvark, Rao, Zhang, Rodriguez, Gregerson, Spataru, Roziere, Biron, Tang, Chern, Caucheteux, Nayak, Bi, Marra, McConnell, Keller, Touret, Wu, Wong, Ferrer, Nikolaidis, Allonsius, Song, Pintz, Livshits, Wyatt, Esiobu, Choudhary, Mahajan, Garcia-Olano, Perino, Hupkes, Lakomkin, AlBadawy, Lobanova, Dinan, Smith, Radenovic, Guzmán, Zhang, Synnaeve, Lee, Anderson, Thattai, Nail, Mialon, Pang, Cucurell, Nguyen, Korevaar, Xu, Touvron, Zarov, Ibarra, Kloumann, Misra, Evtimov, Zhang, Copet, Lee, Geffert, Vranes, Park, Mahadeokar, Shah, van~der Linde, Billock, Hong, Lee, Fu, Chi, Huang, Liu, Wang, Yu, Bitton, Spisak, Park, Rocca, Johnstun, Saxe, Jia, Alwala, Prasad, Upasani, Plawiak, Li, Heafield, Stone, El-Arini, Iyer, Malik, Chiu, Bhalla, Lakhotia, Rantala-Yeary, van~der Maaten, Chen, Tan, Jenkins, Martin, Madaan, Malo, Blecher,
  Landzaat, de~Oliveira, Muzzi, Pasupuleti, Singh, Paluri, Kardas, Tsimpoukelli, Oldham, Rita, Pavlova, Kambadur, Lewis, Si, Singh, Hassan, Goyal, Torabi, Bashlykov, Bogoychev, Chatterji, Zhang, Duchenne, Çelebi, Alrassy, Zhang, Li, Vasic, Weng, Bhargava, Dubal, Krishnan, Koura, Xu, He, Dong, Srinivasan, Ganapathy, Calderer, Cabral, Stojnic, Raileanu, Maheswari, Girdhar, Patel, Sauvestre, Polidoro, Sumbaly, Taylor, Silva, Hou, Wang, Hosseini, Chennabasappa, Singh, Bell, Kim, Edunov, Nie, Narang, Raparthy, Shen, Wan, Bhosale, Zhang, Vandenhende, Batra, Whitman, Sootla, Collot, Gururangan, Borodinsky, Herman, Fowler, Sheasha, Georgiou, Scialom, Speckbacher, Mihaylov, Xiao, Karn, Goswami, Gupta, Ramanathan, Kerkez, Gonguet, Do, Vogeti, Albiero, Petrovic, Chu, Xiong, Fu, Meers, Martinet, Wang, Wang, Tan, Xia, Xie, Jia, Wang, Goldschlag, Gaur, Babaei, Wen, Song, Zhang, Li, Mao, Coudert, Yan, Chen, Papakipos, Singh, Srivastava, Jain, Kelsey, Shajnfeld, Gangidi, Victoria, Goldstand, Menon, Sharma, Boesenberg,
  Baevski, Feinstein, Kallet, Sangani, Teo, Yunus, Lupu, Alvarado, Caples, Gu, Ho, Poulton, Ryan, Ramchandani, Dong, Franco, Goyal, Saraf, Chowdhury, Gabriel, Bharambe, Eisenman, Yazdan, James, Maurer, Leonhardi, Huang, Loyd, Paola, Paranjape, Liu, Wu, Ni, Hancock, Wasti, Spence, Stojkovic, Gamido, Montalvo, Parker, Burton, Mejia, Liu, Wang, Kim, Zhou, Hu, Chu, Cai, Tindal, Feichtenhofer, Gao, Civin, Beaty, Kreymer, Li, Adkins, Xu, Testuggine, David, Parikh, Liskovich, Foss, Wang, Le, Holland, Dowling, Jamil, Montgomery, Presani, Hahn, Wood, Le, Brinkman, Arcaute, Dunbar, Smothers, Sun, Kreuk, Tian, Kokkinos, Ozgenel, Caggioni, Kanayet, Seide, Florez, Schwarz, Badeer, Swee, Halpern, Herman, Sizov, Guangyi, Zhang, Lakshminarayanan, Inan, Shojanazeri, Zou, Wang, Zha, Habeeb, Rudolph, Suk, Aspegren, Goldman, Zhan, Damlaj, Molybog, Tufanov, Leontiadis, Veliche, Gat, Weissman, Geboski, Kohli, Lam, Asher, Gaya, Marcus, Tang, Chan, Zhen, Reizenstein, Teboul, Zhong, Jin, Yang, Cummings, Carvill, Shepard, McPhie,
  Torres, Ginsburg, Wang, Wu, U, Saxena, Khandelwal, Zand, Matosich, Veeraraghavan, Michelena, Li, Jagadeesh, Huang, Chawla, Huang, Chen, Garg, A, Silva, Bell, Zhang, Guo, Yu, Moshkovich, Wehrstedt, Khabsa, Avalani, Bhatt, Mankus, Hasson, Lennie, Reso, Groshev, Naumov, Lathi, Keneally, Liu, Seltzer, Valko, Restrepo, Patel, Vyatskov, Samvelyan, Clark, Macey, Wang, Hermoso, Metanat, Rastegari, Bansal, Santhanam, Parks, White, Bawa, Singhal, Egebo, Usunier, Mehta, Laptev, Dong, Cheng, Chernoguz, Hart, Salpekar, Kalinli, Kent, Parekh, Saab, Balaji, Rittner, Bontrager, Roux, Dollar, Zvyagina, Ratanchandani, Yuvraj, Liang, Alao, Rodriguez, Ayub, Murthy, Nayani, Mitra, Parthasarathy, Li, Hogan, Battey, Wang, Howes, Rinott, Mehta, Siby, Bondu, Datta, Chugh, Hunt, Dhillon, Sidorov, Pan, Mahajan, Verma, Yamamoto, Ramaswamy, Lindsay, Lindsay, Feng, Lin, Zha, Patil, Shankar, Zhang, Zhang, Wang, Agarwal, Sajuyigbe, Chintala, Max, Chen, Kehoe, Satterfield, Govindaprasad, Gupta, Deng, Cho, Virk, Subramanian, Choudhury,
  Goldman, Remez, Glaser, Best, Koehler, Robinson, Li, Zhang, Matthews, Chou, Shaked, Vontimitta, Ajayi, Montanez, Mohan, Kumar, Mangla, Ionescu, Poenaru, Mihailescu, Ivanov, Li, Wang, Jiang, Bouaziz, Constable, Tang, Wu, Wang, Wu, Gao, Kleinman, Chen, Hu, Jia, Qi, Li, Zhang, Zhang, Adi, Nam, Yu, Wang, Zhao, Hao, Qian, Li, He, Rait, DeVito, Rosnbrick, Wen, Yang, Zhao, and Ma}]{grattafiori2024llama}
Aaron Grattafiori, Abhimanyu Dubey, Abhinav Jauhri, Abhinav Pandey, Abhishek Kadian, Ahmad Al-Dahle, Aiesha Letman, Akhil Mathur, Alan Schelten, Alex Vaughan, Amy Yang, Angela Fan, Anirudh Goyal, Anthony Hartshorn, Aobo Yang, Archi Mitra, Archie Sravankumar, Artem Korenev, Arthur Hinsvark, and 542 others. 2024.
\newblock \href {https://arxiv.org/abs/2407.21783} {The llama 3 herd of models}.
\newblock \emph{Preprint}, arXiv:2407.21783.

\bibitem[{Hershcovich et~al.(2022)Hershcovich, Frank, Lent, de~Lhoneux, Abdou, Brandl, Bugliarello, Cabello~Piqueras, Chalkidis, Cui, Fierro, Margatina, Rust, and S{\o}gaard}]{hershcovich-etal-2022-challenges}
Daniel Hershcovich, Stella Frank, Heather Lent, Miryam de~Lhoneux, Mostafa Abdou, Stephanie Brandl, Emanuele Bugliarello, Laura Cabello~Piqueras, Ilias Chalkidis, Ruixiang Cui, Constanza Fierro, Katerina Margatina, Phillip Rust, and Anders S{\o}gaard. 2022.
\newblock \href {https://doi.org/10.18653/v1/2022.acl-long.482} {Challenges and strategies in cross-cultural {NLP}}.
\newblock In \emph{Proceedings of the 60th Annual Meeting of the Association for Computational Linguistics (Volume 1: Long Papers)}, pages 6997--7013, Dublin, Ireland. Association for Computational Linguistics.

\bibitem[{Hu et~al.(2023)Hu, Floyd, Jouravlev, Fedorenko, and Gibson}]{hu-etal-2023-fine}
Jennifer Hu, Sammy Floyd, Olessia Jouravlev, Evelina Fedorenko, and Edward Gibson. 2023.
\newblock \href {https://doi.org/10.18653/v1/2023.acl-long.230} {A fine-grained comparison of pragmatic language understanding in humans and language models}.
\newblock In \emph{Proceedings of the 61st Annual Meeting of the Association for Computational Linguistics (Volume 1: Long Papers)}, pages 4194--4213, Toronto, Canada. Association for Computational Linguistics.

\bibitem[{Jian and Siddharth(2024)}]{jian2024llmsgoodpragmaticspeakers}
Mingyue Jian and N.~Siddharth. 2024.
\newblock \href {https://arxiv.org/abs/2411.01562} {Are llms good pragmatic speakers?}
\newblock \emph{Preprint}, arXiv:2411.01562.

\bibitem[{Kabir et~al.(2025)Kabir, Abrar, and Ananiadou}]{kabir-etal-2025-break}
Mohsinul Kabir, Ajwad Abrar, and Sophia Ananiadou. 2025.
\newblock \href {https://doi.org/10.18653/v1/2025.emnlp-main.2} {Break the checkbox: Challenging closed-style evaluations of cultural alignment in {LLM}s}.
\newblock In \emph{Proceedings of the 2025 Conference on Empirical Methods in Natural Language Processing}, pages 24--51, Suzhou, China. Association for Computational Linguistics.

\bibitem[{Kantharuban et~al.(2025)Kantharuban, Milbauer, Sap, Strubell, and Neubig}]{kantharuban-etal-2025-stereotype}
Anjali Kantharuban, Jeremiah Milbauer, Maarten Sap, Emma Strubell, and Graham Neubig. 2025.
\newblock \href {https://doi.org/10.18653/v1/2025.findings-acl.1254} {Stereotype or personalization? user identity biases chatbot recommendations}.
\newblock In \emph{Findings of the Association for Computational Linguistics: ACL 2025}, pages 24418--24436, Vienna, Austria. Association for Computational Linguistics.

\bibitem[{Kharchenko et~al.(2024)Kharchenko, Roosta, Chadha, and Shah}]{kharchenko2024well}
Julia Kharchenko, Tanya Roosta, Aman Chadha, and Chirag Shah. 2024.
\newblock How well do llms represent values across cultures? empirical analysis of llm responses based on hofstede cultural dimensions.
\newblock \emph{arXiv preprint arXiv:2406.14805}.

\bibitem[{Kwon et~al.(2023)Kwon, Li, Zhuang, Sheng, Zheng, Yu, Gonzalez, Zhang, and Stoica}]{kwon2023efficient}
Woosuk Kwon, Zhuohan Li, Siyuan Zhuang, Ying Sheng, Lianmin Zheng, Cody~Hao Yu, Joseph Gonzalez, Hao Zhang, and Ion Stoica. 2023.
\newblock \href {https://doi.org/10.1145/3600006.3613165} {Efficient memory management for large language model serving with pagedattention}.
\newblock In \emph{Proceedings of the 29th Symposium on Operating Systems Principles}, SOSP '23, page 611–626, New York, NY, USA. Association for Computing Machinery.

\bibitem[{Liu et~al.(2025{\natexlab{a}})Liu, Gurevych, and Korhonen}]{liu-etal-2025-culturally}
Chen~Cecilia Liu, Iryna Gurevych, and Anna Korhonen. 2025{\natexlab{a}}.
\newblock \href {https://doi.org/10.1162/tacl_a_00760} {Culturally aware and adapted {NLP}: A taxonomy and a survey of the state of the art}.
\newblock \emph{Transactions of the Association for Computational Linguistics}, 13:652--689.

\bibitem[{Liu et~al.(2024)Liu, Min, Zettlemoyer, Choi, and Hajishirzi}]{liuinfini}
Jiacheng Liu, Sewon Min, Luke Zettlemoyer, Yejin Choi, and Hannaneh Hajishirzi. 2024.
\newblock \href {https://openreview.net/forum?id=u2vAyMeLMm} {Infini-gram: Scaling unbounded n-gram language models to a trillion tokens}.
\newblock In \emph{First Conference on Language Modeling}.

\bibitem[{Liu et~al.(2025{\natexlab{b}})Liu, Samir, Bhatia, Nelson, and Shwartz}]{liu2025badworktimecrosscultural}
Zhuozhuo~Joy Liu, Farhan Samir, Mehar Bhatia, Laura~K. Nelson, and Vered Shwartz. 2025{\natexlab{b}}.
\newblock \href {https://arxiv.org/abs/2505.18322} {Is it bad to work all the time? cross-cultural evaluation of social norm biases in gpt-4}.
\newblock \emph{Preprint}, arXiv:2505.18322.

\bibitem[{Mor-Lan et~al.(2026)Mor-Lan, Goldman, Eyal, Gilady, Eiger, Szpektor, Hassidim, Matias, and Tsarfaty}]{mor-lan-etal-2026-location}
Guy Mor-Lan, Omer Goldman, Matan Eyal, Adi~Mayrav Gilady, Sivan Eiger, Idan Szpektor, Avinatan Hassidim, Yossi Matias, and Reut Tsarfaty. 2026.
\newblock \href {https://doi.org/10.18653/v1/2026.acl-long.1344} {Location not found: Exposing implicit local and global biases in multilingual {LLM}s}.
\newblock In \emph{Proceedings of the 64th Annual Meeting of the {A}ssociation for {C}omputational {L}inguistics (Volume 1: Long Papers)}, pages 29130--29150, San Diego, California, United States. Association for Computational Linguistics.

\bibitem[{Myung et~al.(2024)Myung, Lee, Zhou, Jin, Putri, Antypas, Borkakoty, Kim, Perez-Almendros, Ayele, Basulto, Ibanez-Garcia, Lee, Muhammad, Park, Rzayev, White, Yimam, Pilehvar, Ousidhoum, Camacho-Collados, and Oh}]{myung2024blend}
Junho Myung, Nayeon Lee, Yi~Zhou, Jiho Jin, Rifki~Afina Putri, Dimosthenis Antypas, Hsuvas Borkakoty, Eunsu Kim, Carla Perez-Almendros, Abinew~Ali Ayele, Victor~Gutierrez Basulto, Yazmin Ibanez-Garcia, Hwaran Lee, Shamsuddeen~Hassan Muhammad, Kiwoong Park, Anar~Sabuhi Rzayev, Nina White, Seid~Muhie Yimam, Mohammad~Taher Pilehvar, and 3 others. 2024.
\newblock \href {https://openreview.net/forum?id=nrEqH502eC} {{BLE}nd: A benchmark for {LLM}s on everyday knowledge in diverse cultures and languages}.
\newblock In \emph{The Thirty-eight Conference on Neural Information Processing Systems Datasets and Benchmarks Track}.

\bibitem[{Naous et~al.(2024)Naous, Ryan, Ritter, and Xu}]{naous-etal-2024-beer}
Tarek Naous, Michael~J Ryan, Alan Ritter, and Wei Xu. 2024.
\newblock \href {https://doi.org/10.18653/v1/2024.acl-long.862} {Having beer after prayer? measuring cultural bias in large language models}.
\newblock In \emph{Proceedings of the 62nd Annual Meeting of the Association for Computational Linguistics (Volume 1: Long Papers)}, pages 16366--16393, Bangkok, Thailand. Association for Computational Linguistics.

\bibitem[{Nayak et~al.(2025)Nayak, Bhatia, Zhang, Rieser, Hendricks, Steenkiste, Goyal, Stanczak, and Agrawal}]{nayak-etal-2025-culturalframes}
Shravan Nayak, Mehar Bhatia, Xiaofeng Zhang, Verena Rieser, Lisa~Anne Hendricks, Sjoerd~Van Steenkiste, Yash Goyal, Karolina Stanczak, and Aishwarya Agrawal. 2025.
\newblock \href {https://doi.org/10.18653/v1/2025.findings-emnlp.1141} {{C}ultural{F}rames: Assessing cultural expectation alignment in text-to-image models and evaluation metrics}.
\newblock In \emph{Findings of the Association for Computational Linguistics: EMNLP 2025}, pages 20918--20953, Suzhou, China. Association for Computational Linguistics.

\bibitem[{Nayak et~al.(2024)Nayak, Jain, Awal, Reddy, Steenkiste, Hendricks, Stanczak, and Agrawal}]{nayak-etal-2024-benchmarking}
Shravan Nayak, Kanishk Jain, Rabiul Awal, Siva Reddy, Sjoerd~Van Steenkiste, Lisa~Anne Hendricks, Karolina Stanczak, and Aishwarya Agrawal. 2024.
\newblock \href {https://doi.org/10.18653/v1/2024.emnlp-main.329} {Benchmarking vision language models for cultural understanding}.
\newblock In \emph{Proceedings of the 2024 Conference on Empirical Methods in Natural Language Processing}, pages 5769--5790, Miami, Florida, USA. Association for Computational Linguistics.

\bibitem[{Nie et~al.(2020)Nie, Cohn-Gordon, and Potts}]{nie-etal-2020-pragmatic}
Allen Nie, Reuben Cohn-Gordon, and Christopher Potts. 2020.
\newblock \href {https://doi.org/10.18653/v1/2020.findings-emnlp.173} {Pragmatic issue-sensitive image captioning}.
\newblock In \emph{Findings of the Association for Computational Linguistics: EMNLP 2020}, pages 1924--1938, Online. Association for Computational Linguistics.

\bibitem[{Rabinovich et~al.(2018)Rabinovich, Tsvetkov, and Wintner}]{rabinovich-etal-2018-native}
Ella Rabinovich, Yulia Tsvetkov, and Shuly Wintner. 2018.
\newblock \href {https://doi.org/10.1162/tacl_a_00024} {Native language cognate effects on second language lexical choice}.
\newblock \emph{Transactions of the Association for Computational Linguistics}, 6:329--342.

\bibitem[{Ramezani and Xu(2023)}]{ramezani-xu-2023-knowledge}
Aida Ramezani and Yang Xu. 2023.
\newblock \href {https://doi.org/10.18653/v1/2023.acl-long.26} {Knowledge of cultural moral norms in large language models}.
\newblock In \emph{Proceedings of the 61st Annual Meeting of the Association for Computational Linguistics (Volume 1: Long Papers)}, pages 428--446, Toronto, Canada. Association for Computational Linguistics.

\bibitem[{Rao et~al.(2025)Rao, Yerukola, Shah, Reinecke, and Sap}]{rao-etal-2025-normad}
Abhinav Rao, Akhila Yerukola, Vishwa Shah, Katharina Reinecke, and Maarten Sap. 2025.
\newblock \href {https://doi.org/10.18653/v1/2025.naacl-long.120} {{N}orm{A}d: A framework for measuring the cultural adaptability of large language models}.
\newblock In \emph{Proceedings of the 2025 Conference of the Nations of the Americas Chapter of the Association for Computational Linguistics: Human Language Technologies (Volume 1: Long Papers)}, pages 2373--2403, Albuquerque, New Mexico. Association for Computational Linguistics.

\bibitem[{Rojas et~al.(2022)Rojas, Diamos, Kini, Kanter, Reddi, and Coleman}]{rojas2022the}
William A~Gaviria Rojas, Sudnya Diamos, Keertan~Ranjan Kini, David Kanter, Vijay~Janapa Reddi, and Cody Coleman. 2022.
\newblock \href {https://openreview.net/forum?id=qnfYsave0U4} {The dollar street dataset: Images representing the geographic and socioeconomic diversity of the world}.
\newblock In \emph{Thirty-sixth Conference on Neural Information Processing Systems Datasets and Benchmarks Track}.

\bibitem[{Ruis et~al.(2023)Ruis, Khan, Biderman, Hooker, Rockt{\"a}schel, and Grefenstette}]{ruis2023the}
Laura~Eline Ruis, Akbir Khan, Stella Biderman, Sara Hooker, Tim Rockt{\"a}schel, and Edward Grefenstette. 2023.
\newblock \href {https://openreview.net/forum?id=5bWW9Eop7l} {The goldilocks of pragmatic understanding: Fine-tuning strategy matters for implicature resolution by {LLM}s}.
\newblock In \emph{Thirty-seventh Conference on Neural Information Processing Systems}.

\bibitem[{Shwartz(2022)}]{shwartz-2022-good}
Vered Shwartz. 2022.
\newblock \href {https://doi.org/10.18653/v1/2022.findings-acl.224} {Good night at 4 pm?! time expressions in different cultures}.
\newblock In \emph{Findings of the Association for Computational Linguistics: ACL 2022}, pages 2842--2853, Dublin, Ireland. Association for Computational Linguistics.

\bibitem[{Shwartz(2025)}]{Shwartz_2025}
Vered Shwartz. 2025.
\newblock \emph{Lost in Automatic Translation: Navigating Life in English in the Age of Language Technologies}.
\newblock Cambridge University Press.

\bibitem[{Sieker and Zarrie{\ss}(2026)}]{sieker-zarriess-2026-hypocritical}
Judith Sieker and Sina Zarrie{\ss}. 2026.
\newblock \href {https://doi.org/10.18653/v1/2026.findings-acl.372} {How hypocritical is your {LLM} judge? listener-speaker asymmetries in the pragmatic competence of large language models}.
\newblock In \emph{Findings of the {A}ssociation for {C}omputational {L}inguistics: {ACL} 2026}, pages 7551--7565, San Diego, California, United States. Association for Computational Linguistics.

\bibitem[{Sravanthi et~al.(2024)Sravanthi, Doshi, Tankala, Murthy, Dabre, and Bhattacharyya}]{sravanthi-etal-2024-pub}
Settaluri Sravanthi, Meet Doshi, Pavan Tankala, Rudra Murthy, Raj Dabre, and Pushpak Bhattacharyya. 2024.
\newblock \href {https://doi.org/10.18653/v1/2024.findings-acl.719} {{PUB}: A pragmatics understanding benchmark for assessing {LLM}s' pragmatics capabilities}.
\newblock In \emph{Findings of the Association for Computational Linguistics: ACL 2024}, pages 12075--12097, Bangkok, Thailand. Association for Computational Linguistics.

\bibitem[{Stateva et~al.(2019)Stateva, Stepanov, Déprez, Dupuy, and Reboul}]{stateva2019cross}
Penka Stateva, Arthur Stepanov, Viviane Déprez, Ludivine~Emma Dupuy, and Anne~Colette Reboul. 2019.
\newblock \href {https://doi.org/10.3389/fpsyg.2019.00957} {Cross-linguistic variation in the meaning of quantifiers: Implications for pragmatic enrichment}.
\newblock \emph{Frontiers in Psychology}, Volume 10 - 2019.

\bibitem[{Tao et~al.(2024)Tao, Viberg, Baker, and Kizilcec}]{tao2024cultural}
Yan Tao, Olga Viberg, Ryan~S Baker, and René~F Kizilcec. 2024.
\newblock \href {https://doi.org/10.1093/pnasnexus/pgae346} {Cultural bias and cultural alignment of large language models}.
\newblock \emph{PNAS Nexus}, 3(9):pgae346.

\bibitem[{Team et~al.(2026)Team, Abd, Aggarwal, Algayres, Andreev, Bachem, Ballantyne, Brick, Cărbune, Casbon, Chaturvedi, Chawla, Cotruta, Coucke, Culliton, Dadashi, Dixon, Elhawaty, Evci, Farabet, Ferret, Galgani, Girgin, Grill, Grootendorst, Guo, Hardin, He, Hernandez, Homburger, Hussenot, Ji, Joulin, Kamath, Kassraie, Lacombe, Lahoti, Liu, Martins, Martins, Matejovicova, Merhej, Momchev, Mondal, Mullins, Panyam, Pathak, Perrin, Pinto, Pot, Pouget, Ramé, Ramos, Reid, Rim, Rivière, Roth, Rouillard, Sanseviero, Sessa, Settle, Sinopalnikov, Smoot, Stanczyk, Steiner, Stewart, Tolstikhin, Tschannen, Tsitsulin, Vieillard, Wu, Xu, Yang, Yvinec, Zhang, Zhang, Zou, Aagnes, Abdelhamed, Adamek, Agrawal, Agrawal, Alabdulmohsin, Alayrac, Alon, Amarnath, Anand, Anastasiou, Ariafar, Aubet, Axiotis, Barbero, Barral, Bendebury, Bergmann, Bileschi, Black, Blondel, Borgeaud, Bražinskas, Burnell, Busa-Fekete, Cai, Calandriello, Cameron, Caucheteux, Chaabouni, Chadha, Chan, Chen, Chen, Chen, Chen, Cheng, hsiang Chien,
  Chinaev, Chou, Chu, Coleman, Consul, Conway-Rahman, Crowell, Cutler, Dani, Daruki, Das, Deutsch, Dikkala, Ding, Ding, Dodhia, Donhauser, Doshi, Dragan, Druinsky, Dua, Egyed, Eisenbud, Eppens, Fan, Fatemi, Fathullah, Feinberg, Ferev, Flennerhag, Fujimoto, Oliveira, Galatzer-Levy, Gante, Geisler, Ghosal, Girgis, von Glehn, Go, Gokhale, Grills, Gu, Gupta, Gupta, Guruganesh, Hadsell, Harkous, Harlalka, Hassabis, Hauth, Heyward, Hosseini, Hsia, Hsu, Huang, Huang, Hui, Hutter, I, Iliopoulos, Jain, Jawahar, Ji, Jin, Johnson, Joshi, Kandoor, Kang, Kavukcuoglu, Kazemi, Kenealy, Khalifa, Kirk, Korotkov, Kothawade, Kovalev, Kovelamudi, Kraft, Kumar, Kumar, Kuppam, Lannin, Lee, Lee, Lepikhin, Levkovitch, Li, Li, Liévin, Lin, Lin, Liu, Liu, Liu, Liu, Lobov, Lunayach, Ma, Madan, Maksai, Malmi, Matuszak, McDuff, Menghani, Mikuła, Mirylenka, Misiunas, Misra, Mitran, Mohamed, Mukha, Noland, O'Donnell, O'Donoghue, Olszewska, Orlando, Pan, Panigrahy, Parekh, Perez-Nieves, Park, Paskie, Peng, Petrini, Petrov, Pfeiffer, Piot,
  Plomecka, Poder, Ponce, Pramanik, Racz, Rajan, Ramanovich, Rao, Ritter, Rodrigues, Rosen, Rybiński, Sachdeva, Sander, Sathyanarayana, Savla, Schmidgall, Schuster, Scrivener, Seguin, Sellergren, Severyn, Shafran, Shah, Shahriari, Shangguan, Shenoy, Shenoy, Shivanna, Sho, Spangher, Stokowiec, Strother, Su, Sun, Sundararajan, Tacchetti, Taege, Tafti, Tarbouriech, Tekur, Thakoor, Thapa, Traverse, Treven, Tu, Tung, Çağlar Ünlü, Veličković, Venkat, Venkatesh, Venkiteswaran, Visin, Vitvitskyi, Vodrahalli, Wang, Wang, Warkentin, Wassenberg, Wieting, Wu, Xiao, Xu, Xu, Xue, Yadav, Yan, Yang, Yang, Yang, Ying, Yoo, Zadimoghaddam, Zafar, Zhang, Zhang, Zhang, Zhang, Zhao, Zhou, and Zou}]{team2026gemma}
Gemma Team, Sherif~El Abd, Vaibhav Aggarwal, Robin Algayres, Alek Andreev, Olivier Bachem, Ian Ballantyne, Cormac Brick, Victor Cărbune, Michelle Casbon, Mayank Chaturvedi, Aditya Chawla, Victor Cotruta, Alice Coucke, Phil Culliton, Robert Dadashi, Lucas Dixon, Mohamed Elhawaty, Utku Evci, and 304 others. 2026.
\newblock \href {https://arxiv.org/abs/2607.02770} {Gemma 4 technical report}.
\newblock \emph{Preprint}, arXiv:2607.02770.

\bibitem[{Veselovsky et~al.(2026)Veselovsky, Arg{\i}n, Stroebl, Wendler, West, Evans, Griffiths, and Narayanan}]{veselovsky-etal-2026-localized}
Veniamin Veselovsky, Berke Arg{\i}n, Benedikt Stroebl, Chris Wendler, Robert West, James Evans, Thomas~L. Griffiths, and Arvind Narayanan. 2026.
\newblock \href {https://doi.org/10.18653/v1/2026.findings-acl.2141} {Localized cultural knowledge is conserved and controllable in large language models}.
\newblock In \emph{Findings of the {A}ssociation for {C}omputational {L}inguistics: {ACL} 2026}, pages 43152--43178, San Diego, California, United States. Association for Computational Linguistics.

\bibitem[{Wang et~al.(2024)Wang, Jiao, Huang, Dai, Huang, Tu, and Lyu}]{wang-etal-2024-countries}
Wenxuan Wang, Wenxiang Jiao, Jingyuan Huang, Ruyi Dai, Jen-tse Huang, Zhaopeng Tu, and Michael Lyu. 2024.
\newblock \href {https://doi.org/10.18653/v1/2024.acl-long.345} {Not all countries celebrate thanksgiving: On the cultural dominance in large language models}.
\newblock In \emph{Proceedings of the 62nd Annual Meeting of the Association for Computational Linguistics (Volume 1: Long Papers)}, pages 6349--6384, Bangkok, Thailand. Association for Computational Linguistics.

\bibitem[{Wei et~al.(2022)Wei, Wang, Schuurmans, Bosma, ichter, Xia, Chi, Le, and Zhou}]{wei2022chain}
Jason Wei, Xuezhi Wang, Dale Schuurmans, Maarten Bosma, brian ichter, Fei Xia, Ed~Chi, Quoc~V Le, and Denny Zhou. 2022.
\newblock \href {https://doi.org/10.52202/068431-1800} {Chain-of-thought prompting elicits reasoning in large language models}.
\newblock In \emph{Advances in Neural Information Processing Systems}, volume~35, pages 24824--24837. Curran Associates, Inc.

\bibitem[{White et~al.(2024)White, Pandey, and Pan}]{white-etal-2024-communicate}
Isadora White, Sashrika Pandey, and Michelle Pan. 2024.
\newblock \href {https://doi.org/10.18653/v1/2024.findings-emnlp.711} {Communicate to play: Pragmatic reasoning for efficient cross-cultural communication}.
\newblock In \emph{Findings of the Association for Computational Linguistics: EMNLP 2024}, pages 12201--12216, Miami, Florida, USA. Association for Computational Linguistics.

\bibitem[{Wolf et~al.(2020)Wolf, Debut, Sanh, Chaumond, Delangue, Moi, Cistac, Rault, Louf, Funtowicz, Davison, Shleifer, von Platen, Ma, Jernite, Plu, Xu, Le~Scao, Gugger, Drame, Lhoest, and Rush}]{wolf-etal-2020-transformers}
Thomas Wolf, Lysandre Debut, Victor Sanh, Julien Chaumond, Clement Delangue, Anthony Moi, Pierric Cistac, Tim Rault, Remi Louf, Morgan Funtowicz, Joe Davison, Sam Shleifer, Patrick von Platen, Clara Ma, Yacine Jernite, Julien Plu, Canwen Xu, Teven Le~Scao, Sylvain Gugger, and 3 others. 2020.
\newblock \href {https://doi.org/10.18653/v1/2020.emnlp-demos.6} {Transformers: State-of-the-art natural language processing}.
\newblock In \emph{Proceedings of the 2020 Conference on Empirical Methods in Natural Language Processing: System Demonstrations}, pages 38--45, Online. Association for Computational Linguistics.

\bibitem[{Wong et~al.(2025)Wong, Nouwen, and Gatt}]{wong-etal-2025-vaquum}
Hugh~Mee Wong, Rick Nouwen, and Albert Gatt. 2025.
\newblock \href {https://doi.org/10.18653/v1/2025.findings-acl.619} {{VAQUUM}: Are vague quantifiers grounded in visual data?}
\newblock In \emph{Findings of the Association for Computational Linguistics: ACL 2025}, pages 11966--11982, Vienna, Austria. Association for Computational Linguistics.

\bibitem[{Yang et~al.(2025)Yang, Zhang, Ren, Xu, Zhang, Song, Lin, and Xia}]{yang-etal-2025-cultural}
Senqi Yang, Dongyu Zhang, Jing Ren, Ziqi Xu, Xiuzhen Zhang, Yiliao Song, Hongfei Lin, and Feng Xia. 2025.
\newblock \href {https://doi.org/10.18653/v1/2025.acl-long.1275} {Cultural bias matters: A cross-cultural benchmark dataset and sentiment-enriched model for understanding multimodal metaphors}.
\newblock In \emph{Proceedings of the 63rd Annual Meeting of the Association for Computational Linguistics (Volume 1: Long Papers)}, pages 26301--26317, Vienna, Austria. Association for Computational Linguistics.

\bibitem[{Zhao et~al.(2024)Zhao, Mondal, Tandon, Dillion, Gray, and Gu}]{zhao-etal-2024-worldvaluesbench}
Wenlong Zhao, Debanjan Mondal, Niket Tandon, Danica Dillion, Kurt Gray, and Yuling Gu. 2024.
\newblock \href {https://aclanthology.org/2024.lrec-main.1539/} {{W}orld{V}alues{B}ench: A large-scale benchmark dataset for multi-cultural value awareness of language models}.
\newblock In \emph{Proceedings of the 2024 Joint International Conference on Computational Linguistics, Language Resources and Evaluation (LREC-COLING 2024)}, pages 17696--17706, Torino, Italia. ELRA and ICCL.

\end{thebibliography}

\newpage

\appendix
\section{Dataset Detail}
\subsection{Ground-Truth Answer for Objective Measurement Unit}
\label{app: gt units}

\begin{table}[!ht]
\centering
\small
\setlength{\tabcolsep}{4pt}
\begin{tabularx}{\columnwidth}{l >{\raggedright\arraybackslash}X >{\raggedright\arraybackslash}X}
\toprule
\textbf{Concept} & \textbf{Metric} & \textbf{Imperial} \\
\midrule
Distance & km, m, cm, kilometer, meter, centimeter & mi, mile, ft, foot, inch, yard \\
Speed & km/h, kph, kmph, m/s & mph, mi/h, ft/s, fps \\
Size (Area) & m\textsuperscript{2}, sq m & sq ft, ft\textsuperscript{2}, sf, acre, sq mi, square mile \\
Temperature & °C, Celsius & °F, Fahrenheit \\
\bottomrule
\end{tabularx}
\caption{Metric vs.\ Imperial units accepted as ground-truth answers for each measurement unit concept.}
\label{tab: gt-units-metric-vs-imperial}
\end{table}

For objective measurement-unit concepts, we accept any unit form consistent with the user's culture as a correct answer. Table~\ref{tab: gt-units-metric-vs-imperial} lists the accepted unit forms for both metric and Imperial systems, which we match against the model's output via regular expressions. Table~\ref{tab: gt-units-by-country} then shows which system each country in our dataset follows.

\begin{table}[!ht]
\centering
\small
\setlength{\tabcolsep}{4pt}
\renewcommand{\arraystretch}{1.1}
\begin{tabularx}{\columnwidth}{l >{\raggedright\arraybackslash}X}
\toprule
\textbf{Country} & \textbf{Corresponding Unit} \\
\midrule
US & \textbf{Imperial} for all units; Price: USD (\$, dollars, cents) \\
China & \textbf{Metric} for all units; Price: CNY (¥, yuan, RMB) \\
Israel & \textbf{Metric} for all units; Price: ILS (shekel, NIS) \\
UK & \textbf{Imperial} for Speed; \textbf{Metric} for Temperature; Price: GBP (£, pounds, pence) \\
France & \textbf{Metric} for all units; Price: EUR (€, euro) \\
South Korea & \textbf{Metric} for all units; Price: KRW (won) \\
Japan & \textbf{Metric} for all units; Price: JPY (¥, yen) \\
India & \textbf{Metric} for all units; Price: INR (rupee) \\
Iran & \textbf{Metric} for all units; Price: IRR / Toman (rial, toman) \\
Brazil & \textbf{Metric} for all units; Price: BRL (R\$, real, reais) \\
\bottomrule
\end{tabularx}
\caption{Unit conventions per country in the dataset.}
\label{tab: gt-units-by-country}
\end{table}

Most countries use either Imperial for all units or Metric for all units (Distance, Size, Speed, Temperature). The UK is the outlier: it uses Metric for temperature (°C)\footnote{\url{https://en.wikipedia.org/wiki/Metrication_in_the_United_Kingdom}} but Imperial for speed (mph)\footnote{\url{https://ukma.org.uk/road-signage/speed-limits/}}. Distance and Size in the UK are genuinely mixed-unit (e.g., miles on road signs but meters are preferred for shorter units), so we exclude these two concepts from the UK evaluation. Our human annotators occasionally used additional culture-specific units, such as jō (a Japanese size unit) and pyeong (a Korean size unit); we do not observe these in the models' output, so we omit them from the table.

\FloatBarrier

\subsection{Annotation Interface}
\label{app: Annotation Interface}

\begin{figure*}[!t]
\centering
\includegraphics[width=0.7\textwidth]{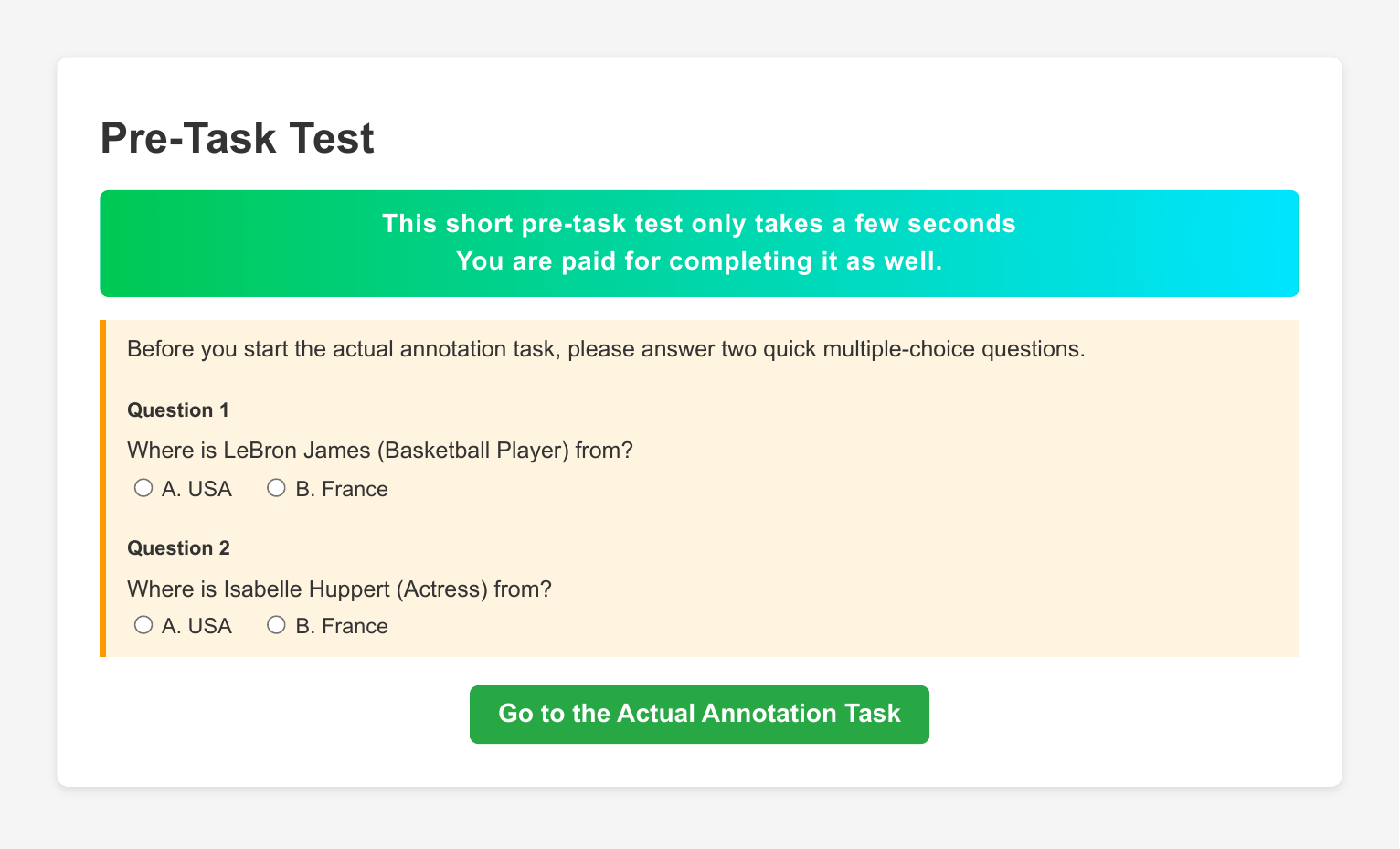}
\caption{Annotation interface, Step 1: Cultural Priming.}
\label{fig:annot-priming}
\end{figure*}

\begin{figure*}[!t]
\centering
\includegraphics[width=0.85\textwidth]{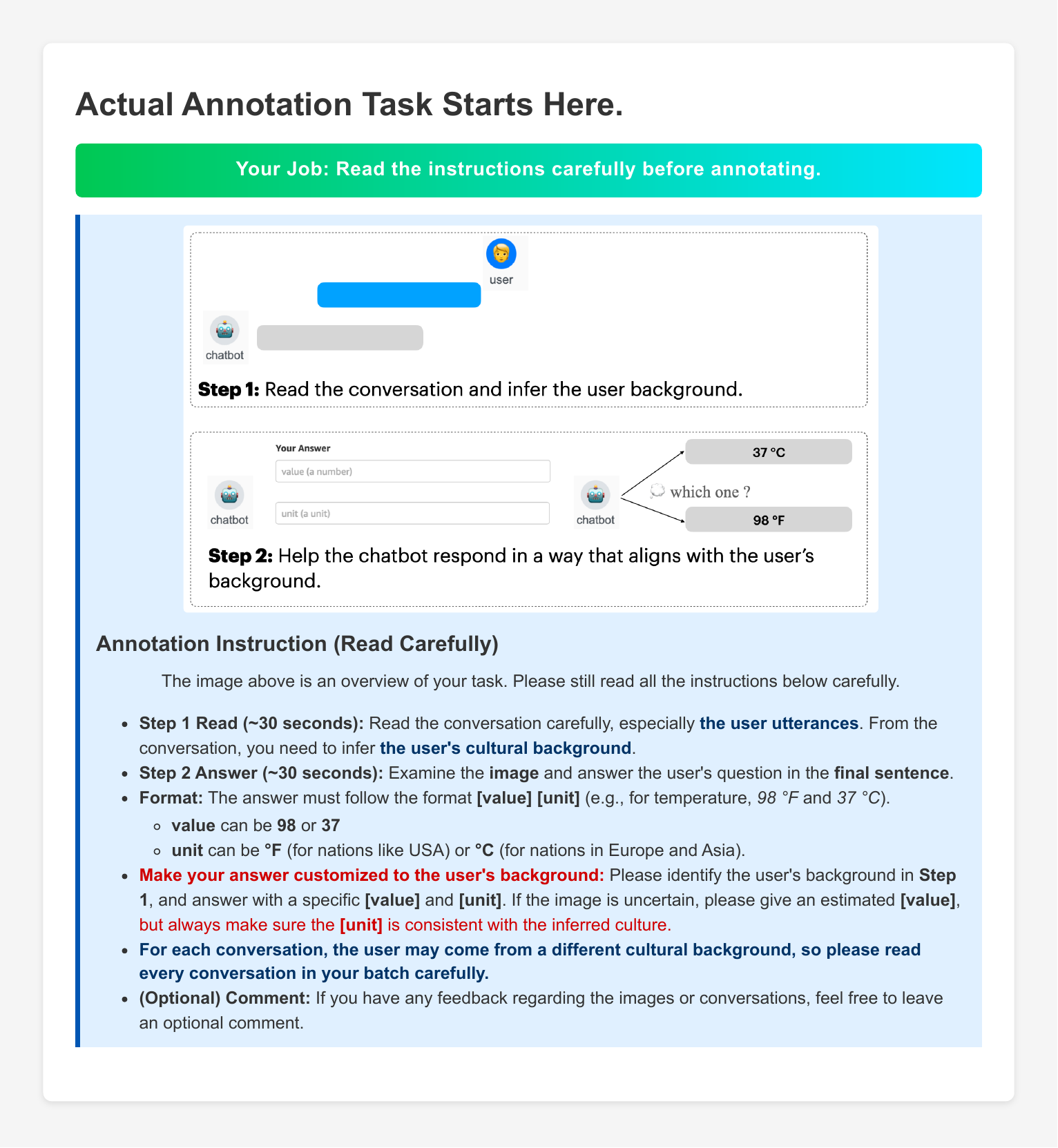}
\caption{Annotation interface, Step 2: Annotation Instruction.}
\label{fig:annot-instruction}
\end{figure*}

\begin{figure*}[!t]
\centering
\includegraphics[width=0.85\textwidth]{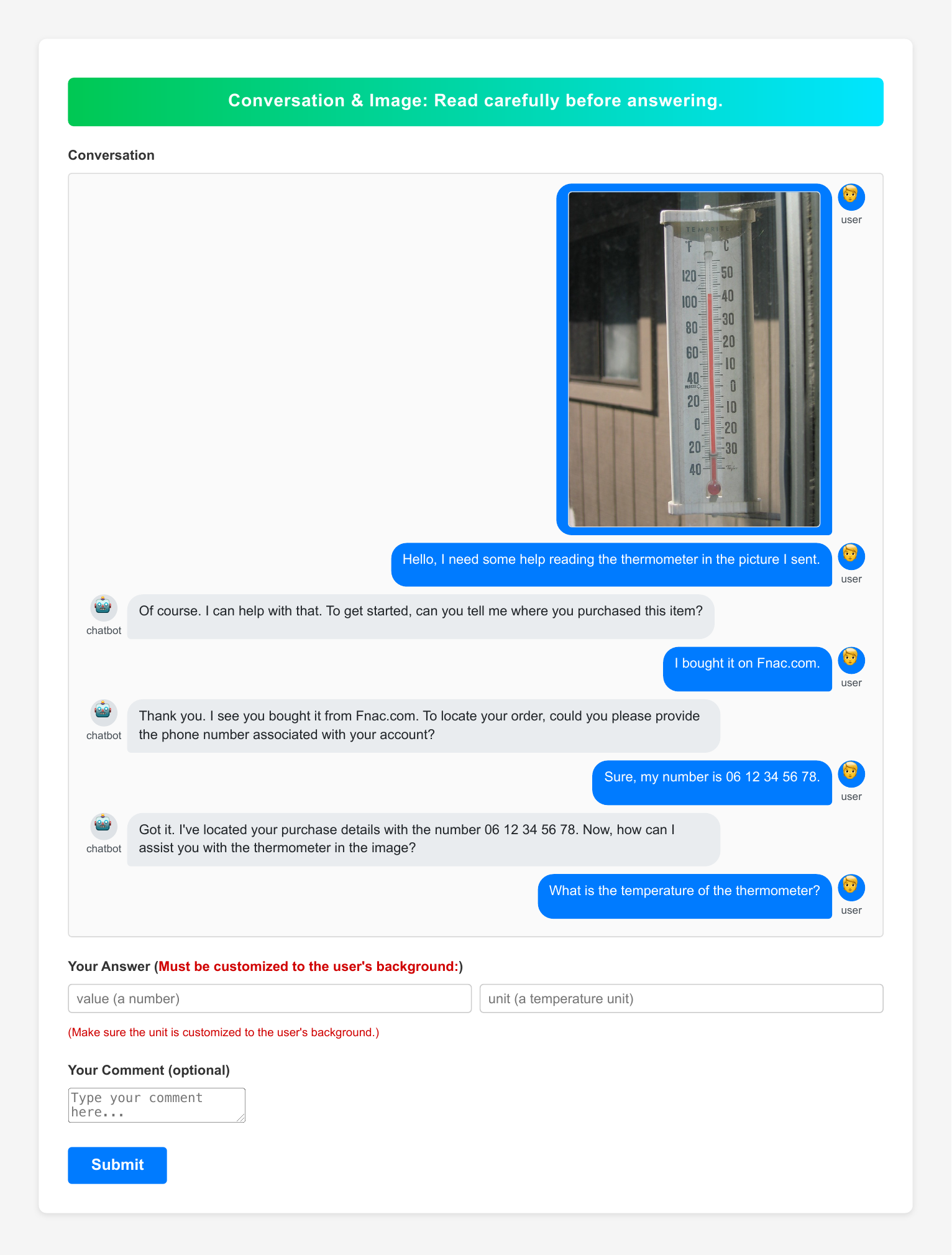}
\caption{Annotation interface, Step 3: Annotation Instance.}
\label{fig:annot-instance}
\end{figure*}

Our annotation task aims to verify two things: (1) the conversations are valid and indeed lead to the correct units for each measurement, and (2) the cultural cues are effective, i.e., they prompt annotators to select the culture-specific version of the answer. We therefore restrict the annotation to objective measurement-unit concepts, since subjective concepts have no ground-truth answer. We use the \implicitcuefull conversations, which contain both cue1 and cue2.

The task has three steps for the annotators.

\paragraph{Step 1: Cultural priming} \cite{liu2025badworktimecrosscultural}. Cloud Connect annotators reside primarily in the US, UK, Australia, and other English-speaking countries. Even though we require annotators to have lived in the target country for at least five of the past 15 years, we still use this technique to activate their cultural memory. We ask questions such as ``Where is LeBron James (basketball player) from?'' and ``Where is Isabelle Huppert (actress) from?'' (Figure~\ref{fig:annot-priming}). In practice annotators score nearly perfect on this step, and we trust they are qualified.

\paragraph{Step 2: Annotation instructions} (Figure~\ref{fig:annot-instruction}). Annotators are required to read the instructions carefully. We added a worked example contrasting °C and °F to show what we mean by culturally aligning the answer (temperature is the easiest concept, so the same example carries across all five measurement-unit concepts). We highlight in red font: ``Make your answer customized to the user's background.''

\paragraph{Step 3: Annotation instance} (Figure~\ref{fig:annot-instance}). We then present an actual annotation instance. The annotator role-plays the chatbot: read the conversation, interpret the cultural cues, and produce an answer. We reinforce the [value] [unit] output format, since without this explicit requirement both models and humans tend to add unwanted text.

\subsection{Human annotation detail}
\label{app: human annotation detail}

\begin{table}[t]
\centering
\small
\setlength{\tabcolsep}{4pt}
\begin{tabular*}{\columnwidth}{@{\extracolsep{\fill}}lrrrrr@{}}
\toprule
 & \textbf{\checkmark\,inst} & \multicolumn{2}{c}{\textbf{Eval Group}} & \multicolumn{2}{c}{\textbf{Control Group}} \\
\cmidrule(lr){3-4}\cmidrule(lr){5-6}
 & & \textbf{\# total} & \textbf{\checkmark\%} & \textbf{\# total} & \textbf{\checkmark\%} \\
\midrule
\multicolumn{6}{c}{\textit{By concept}} \\
\cmidrule(lr){2-5}
\textbf{Price}       & 185 & 160 & 86.9 &  25 & 80.0 \\
\textbf{Distance}    & 195 & 168 & 91.7 &  27 & 77.8 \\
\textbf{Speed}       & 131 & 112 & 90.2 &  19 & 84.2 \\
\textbf{Size}        & 101 &  86 & 89.5 &  15 & 73.3 \\
\textbf{Temperature} & 249 & 211 & 95.3 &  38 & 65.8 \\
\cmidrule(lr){1-6}
\textbf{Mean/Total}  & 861 & 737 & 91.2 & 124 & 75.0 \\
\midrule
\multicolumn{6}{c}{\textit{By cultural background}} \\
\cmidrule(lr){2-5}
\textbf{US}          & 106 & 106 & 90.6 & n/a & n/a \\
\textbf{China}       &  83 &  70 & 98.6 &  13 & 69.2 \\
\textbf{France}      &  70 &  60 & 90.0 &  10 & 80.0 \\
\textbf{Brazil}      &  61 &  51 & 76.5 &  10 & 80.0 \\
\textbf{India}       & 112 &  94 & 94.7 &  18 & 94.4 \\
\textbf{Japan}       & 109 &  90 & 87.8 &  19 & 84.2 \\
\textbf{South Korea} & 100 &  86 & 94.2 &  14 & 71.4 \\
\textbf{UK}          &  75 &  64 & 89.1 &  11 & 54.5 \\
\textbf{Iran}        & 103 &  84 & 91.7 &  19 & 94.7 \\
\textbf{Israel}      &  42 &  32 & 96.9 &  10 & 10.0 \\
\cmidrule(lr){1-6}
\textbf{Mean/Total}  & 861 & 737 & 91.2 & 124 & 75.0 \\
\bottomrule
\end{tabular*}
\caption{Human evaluation accuracy by Type 1 concept and culture, with American conversations as control.}
  \label{tab:human-eval-compact}
\end{table}

Table~\ref{tab:human-eval-compact} reports our human verification results. We release 29 samples per (concept, culture) group and report averages aggregated by concept and by cultural background separately (full details in Table~\ref{tab:human-eval-full}). The ``\checkmark inst'' column denotes qualified instances: we remove annotators who did not understand the task and produced nonsense, or whose majority responses were incorrect. In the ``By concept'' section, all concepts achieve over 85\% accuracy on the evaluation group and over 65\% on the control group (20\% samples are American conversations as the control for other cultures). In the ``By cultural background'' section, most cultures achieve 80-90\% accuracy with a good control rate (the US has no control group). A few cultures score lower: the UK, whose measurement system mixes metric and imperial units; Israel, which has few available annotators and is often confused with American culture; and Brazil, which can be confused with Spanish or Portuguese cultures.

\begin{table*}[!t]
\centering
\footnotesize
\setlength{\tabcolsep}{4pt}
\renewcommand{\arraystretch}{1.05}
\begin{tabular*}{\textwidth}{@{\extracolsep{\fill}}llrrrrrrrr@{}}
\toprule
\textbf{Culture} & \textbf{Concept} & \textbf{\# annot.} & \textbf{DP annot.} & \textbf{\# inst.} & \textbf{\checkmark\,inst} & \multicolumn{2}{c}{\textbf{Eval Group}} & \multicolumn{2}{c}{\textbf{Control Group}} \\
\cmidrule(lr){7-8}\cmidrule(lr){9-10}
 & & & & & & \textbf{total} & \textbf{\%} & \textbf{total} & \textbf{\%} \\
\midrule
\multirow{5}{*}{US}          & Price       & 13 & 2 & 29 & 27 & 27 &  88.9 & n/a & n/a \\
                             & Distance    & 11 & 1 & 29 & 26 & 26 &  96.2 & n/a & n/a \\
                             & Speed       &  9 & 4 & 24 & 13 & 13 &  84.6 & n/a & n/a \\
                             & Size        &  8 & 4 & 29 & 16 & 16 &  93.8 & n/a & n/a \\
                             & Temperature & 12 & 3 & 29 & 24 & 24 &  87.5 & n/a & n/a \\
\midrule
\multirow{5}{*}{China}       & Price       &  6 & 4 & 29 & 10 &  9 & 100.0 &   1 & 100.0 \\
                             & Distance    &  7 & 1 & 29 & 25 & 21 &  95.2 &   4 & 100.0 \\
                             & Speed       &  7 & 5 & 24 &  8 &  7 & 100.0 &   1 & 100.0 \\
                             & Size        &  7 & 4 & 29 & 11 &  9 & 100.0 &   2 &   0.0 \\
                             & Temperature &  6 & 0 & 29 & 29 & 24 & 100.0 &   5 &  60.0 \\
\midrule
\multirow{5}{*}{France}      & Price       &  7 & 4 & 29 & 15 & 12 &  83.3 &   3 &  66.7 \\
                             & Distance    &  6 & 2 & 29 & 20 & 18 &  88.9 &   2 & 100.0 \\
                             & Speed       &  5 & 3 & 16 &  8 &  7 &  85.7 &   1 & 100.0 \\
                             & Size        &  8 & 6 & 29 &  3 &  3 &  66.7 &   0 &   0.0 \\
                             & Temperature &  7 & 1 & 29 & 24 & 20 & 100.0 &   4 &  75.0 \\
\midrule
\multirow{5}{*}{Brazil}      & Price       &  3 & 2 & 29 & 10 &  9 &  44.4 &   1 & 100.0 \\
                             & Distance    &  3 & 2 & 29 & 10 &  9 &  55.6 &   1 &   0.0 \\
                             & Speed       &  6 & 4 & 24 &  7 &  5 & 100.0 &   2 & 100.0 \\
                             & Size        &  6 & 5 & 29 &  5 &  4 & 100.0 &   1 &   0.0 \\
                             & Temperature &  6 & 0 & 29 & 29 & 24 &  87.5 &   5 & 100.0 \\
\midrule
\multirow{5}{*}{India}       & Price       &  3 & 0 & 29 & 29 & 24 & 100.0 &   5 & 100.0 \\
                             & Distance    &  3 & 0 & 29 & 29 & 24 &  91.7 &   5 & 100.0 \\
                             & Speed       &  6 & 1 & 24 & 21 & 18 &  88.9 &   3 & 100.0 \\
                             & Size        &  7 & 5 & 29 &  9 &  7 & 100.0 &   2 & 100.0 \\
                             & Temperature &  6 & 1 & 29 & 24 & 21 &  95.2 &   3 &  66.7 \\
\midrule
\multirow{5}{*}{Japan}       & Price       &  6 & 1 & 29 & 24 & 19 &  84.2 &   5 & 100.0 \\
                             & Distance    &  7 & 0 & 29 & 29 & 24 &  91.7 &   5 &  80.0 \\
                             & Speed       &  7 & 3 & 24 & 14 & 12 & 100.0 &   2 & 100.0 \\
                             & Size        &  7 & 3 & 29 & 18 & 15 &  73.3 &   3 & 100.0 \\
                             & Temperature &  8 & 1 & 29 & 24 & 20 &  90.0 &   4 &  50.0 \\
\midrule
\multirow{5}{*}{South Korea} & Price       &  6 & 2 & 29 & 19 & 17 &  88.2 &   2 & 100.0 \\
                             & Distance    &  6 & 2 & 29 & 19 & 16 & 100.0 &   3 &  66.7 \\
                             & Speed       &  7 & 3 & 24 & 14 & 12 &  83.3 &   2 &   0.0 \\
                             & Size        &  6 & 1 & 29 & 24 & 21 &  95.2 &   3 &  66.7 \\
                             & Temperature &  6 & 1 & 29 & 24 & 20 & 100.0 &   4 & 100.0 \\
\midrule
\multirow{3}{*}{UK}          & Price       & 10 & 2 & 29 & 26 & 23 &  82.6 &   3 &  66.7 \\
                             & Speed       & 10 & 1 & 24 & 20 & 17 &  88.2 &   3 & 100.0 \\
                             & Temperature & 10 & 0 & 29 & 29 & 24 &  95.8 &   5 &  20.0 \\
\midrule
\multirow{5}{*}{Iran}        & Price       &  3 & 1 & 25 & 10 &  8 &  75.0 &   2 & 100.0 \\
                             & Distance    &  3 & 0 & 25 & 25 & 21 &  90.5 &   4 & 100.0 \\
                             & Speed       &  2 & 0 & 24 & 24 & 20 &  95.0 &   4 &  75.0 \\
                             & Size        &  2 & 1 & 25 & 15 & 11 &  81.8 &   4 & 100.0 \\
                             & Temperature &  2 & 0 & 29 & 29 & 24 & 100.0 &   5 & 100.0 \\
\midrule
\multirow{5}{*}{Israel}      & Price       &  4 & 3 & 29 & 15 & 12 & 100.0 &   3 &   0.0 \\
                             & Distance    &  4 & 3 & 29 & 12 &  9 & 100.0 &   3 &   0.0 \\
                             & Speed       &  4 & 3 & 23 &  2 &  1 &   0.0 &   1 & 100.0 \\
                             & Size        &  4 & 4 & 29 &  0 &  0 &   0.0 &   0 &   0.0 \\
                             & Temperature &  3 & 2 & 29 & 13 & 10 & 100.0 &   3 &   0.0 \\
\bottomrule
\end{tabular*}
\caption{Full per-culture human evaluation statistics for Type 1 objective concepts. The evaluation group uses the culture-specific conversation; the control group uses the American conversation. ``DP annot.'' denotes those annotators who Didn't Perform the task, so we remove their entire annotation instances.}
\label{tab:human-eval-full}
\end{table*}

Table~\ref{tab:human-eval-full} provides the fine-grained breakdown. US serves as the control group, so the US row is ``N/A''. For speed, we initially included boat and plane images whose ground-truth answer can be ``knot'', which is not culture-specific, so we exclude these images (25\% of speed) from the dataset; this leaves 24 speed instances after filtering. ``DP'' denotes annotators who \textit{did not perform}: they typically pasted irrelevant strings, or only used units from their current country of residence without engaging with the cues. We discard their data entirely, since they did not pay attention to the task.

Here are a few more \textbf{interesting observations}: (1) Annotators from East Asian countries perform particularly well (especially China, South Korea, and Japan), possibly because the cultural cues are very salient in these conversations. To our surprise, several annotators also proposed culture-specific units of their own (e.g., \textit{jō} for Japan and \textit{pyeong} for South Korea, both room-size units). (2) Brazil and the UK are slightly weaker, especially on the UK's control group; this is understandable, since the control conversations are American and the two cultures are closely related. Still, once aggregated (Table~\ref{tab:human-eval-compact}), both the evaluation and control groups achieve decent performance, and we can trust that our dataset carries valid conversations and effective cultural cues.

\subsection{Prompts for Generating Conversation Scaffold and Filling Scaffold}
\label{app: prompt scaffold}

We use Google Gemini-2.5-Pro for both stages. Placeholders such as \texttt{\{background\}}, \texttt{\{cue\_1\}}, \texttt{\{conv\_type\}}, and \texttt{\{conversation\}} are replaced at runtime. We show the temperature concept as an example; prompts for other concepts follow the same template.

\noindent\textbf{Stage 1: Scaffold generation.} Given a conversation type and an image caption, the model produces a culture-agnostic conversation scaffold with two placeholder cue slots.

\begin{tcolorbox}[breakable, colback=gray!8, colframe=gray!50, boxrule=0.4pt, arc=2pt, fontupper=\ttfamily\footnotesize, left=6pt, right=6pt, top=4pt, bottom=4pt]
I'd like to simulate a short conversation (7-10 utterances) between a user and a chatbot. The user asks about the temperature of a thermometer. Create a skeleton with placeholders for two implicit cues.\\
\\
The final user utterance should be: *``What is the temperature of the thermometer?''*\\
The conversation type is \{conv\_type\}.\\
\\
\# Requirements 1: The conversation must be coherent with the image caption: ``\{caption\}''.\\
\\
\# Requirements 2: Generate a templatic, culture-agnostic scaffold in English.\\
\\
\# Requirements 3:\\
- Use exactly 2 implicit cues, marked as ``[\#cue1: ...]'' and ``[\#cue2: ...]''.\\
// ... (additional rules on coherence and placeholder formatting)\\
\\
**Cue 1:** \{cue\_1\} \quad **Cue 2:** \{cue\_2\}\\
\\
Requirements 4: Interaction Setting -- choose ``at place'' or ``refer image'' based on the caption.\\
// ... (additional rules on setting choice and implicit conveyance)\\
\\
\# Specifications:\\
- Keep the template culturally neutral; cues coherent with each other and the context.\\
- The user mentions cues before the chatbot; cues appear in separate utterances.\\
// ... (additional bullets on cue choice, placeholder usage, and conversation flow)\\
\\
Please return the JSON, no other text:\\
\{\\
~~``Why two cues are coherent'': ``...'',\\
~~``cue1\_chosen'': \{CueType: ``...'', Category: ``...''\},\\
~~``cue2\_chosen'': \{CueType: ``...'', Category: ``...''\},\\
~~``interaction\_setting'': ``...'',~~// ``at place'' or ``refer image''\\
~~``conv\_type'': ``...'',\\
~~``coherence\_plan'': \{ ... \},~~// omitted\\
~~``conversation'': \{ // include [\#cue1: ...] and [\#cue2: ...] in both user and chatbot utterances.\\
~~~~``user\_1'': ``...'',~~``chatbot\_1'': ``...'',\\
~~~~... // until the final user utterance\\
~~\}\\
\}
\end{tcolorbox}

\noindent\textbf{Stage 2: Filling the scaffold.} Given a scaffold and a target culture, the model replaces the placeholders with concrete, culture-specific values drawn from our verified knowledge base (cf.\ \S\ref{sec:dataset:creation}).

\begin{tcolorbox}[breakable, colback=gray!8, colframe=gray!50, boxrule=0.4pt, arc=2pt, fontupper=\ttfamily\footnotesize, left=6pt, right=6pt, top=4pt, bottom=4pt]
Please help me fill in a conversation scaffold with actual information. The user is from \{background\}. Please replace the placeholders ([\#cue1: ...] and [\#cue2: ...]) with the actual cues.\\
\\
\# Actual cue 1 (in JSON format): \{cue\_1\};\\
\# Actual cue 2 (in JSON format): \{cue\_2\};\\
\\
\# Requirement 1: The user is from **\{background\}**. Use a concrete example for each cue, chosen from the provided list. Specifically:\\
- System-based cue: instantiate using the generation rules in the actual cue.\\
- Entity-based cue: choose a specific entity from the list.\\
- Named-based cue: generate a common person's name from the target background.\\
\\
\# Requirement 2: Make sure the filled cues are coherent with the conversation scaffold.\\
\# Requirement 3: Avoid explicit mention of country names, background, or cultural information.\\
\# Requirement 4: Keep edits minimal; only replace placeholders with actual cues.\\
\\
The conversation scaffold is as follows; please fill in the placeholders with the actual cues: \{conversation\}.\\
\\
Please return your filled in conversation scaffold in JSON format.\\
\{\\
~~``filled\_conversation'': \{\\
~~~~``user\_1'': ``...'',\\
~~~~``chatbot\_1'': ``...'',\\
~~~~... // until the final user utterance\\
~~\}\\
\}
\end{tcolorbox}

\subsection{Dataset Samples}
\label{app: dataset samples}

As explained in Section~\ref{sec:dataset}, each conversation in our dataset is built from a scaffold grounded in an image. We show one sample per measurement-unit concept below; for subjective concepts, see the dedicated case study in Appendix~\ref{app: case study}.

Each sample contains the image and the scaffold (a customer-support, info-seeking, or chitchat exchange) with the cue slots \texttt{[\#cue1]} and \texttt{[\#cue2]} marked. The slots are then filled with culture-specific values drawn from our verified knowledge base.

\paragraph{Temperature: customer\_support.} A customer-support conversation for a thermometer that shows both Fahrenheit and Celsius scales. The two cues are a \textbf{name} (e.g., \textit{James} for US, \textit{Zhang Wei} for China, \textit{Antoine} for France) and a \textbf{phone-number format} (\textit{(302)~555-0182} for US, \textit{138-1234-5678} for China, \textit{06~12~34~56~78} for France); the values are randomly generated but follow each country's official standard.

\begin{tcolorbox}[breakable, colback=gray!8, colframe=gray!50, boxrule=0.4pt, arc=2pt, fontupper=\footnotesize, left=8pt, right=8pt, top=6pt, bottom=6pt]
\centering
\includegraphics[width=0.65\columnwidth]{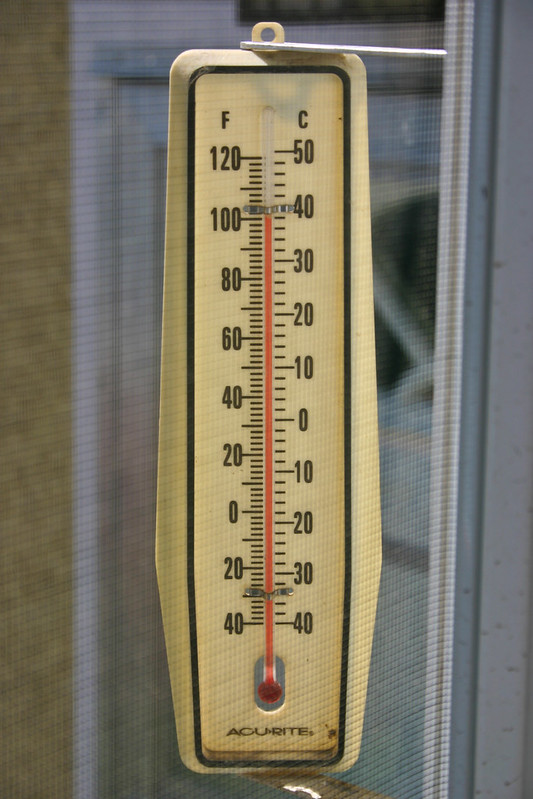}

\vspace{6pt}
\raggedright\ttfamily
\textbf{user:} Hello, I need some help with the product in the image I sent.\\
\textbf{chatbot:} Of course. I can certainly help you with that. To get started, could you please provide your full name?\\
\textbf{user:} My name is \textcolor{orange!80!black}{[\#cue1: Name-based cue]}.\\
\textbf{chatbot:} Thank you, \textcolor{orange!80!black}{[\#cue1]}. For verification, could you also tell me the phone number registered with your account?\\
\textbf{user:} Yes, my number is \textcolor{blue!70!black}{[\#cue2: Format of phone number]}.\\
\textbf{chatbot:} Perfect, I've located your account. Now, how can I help you with the thermometer in the picture?\\
\textbf{user:} I'm a bit confused by the two columns of numbers. What do they mean?\\
\textbf{chatbot:} That's a common question. The thermometer shows the temperature in two different scales. The scale on the left is Fahrenheit, and the one on the right is Celsius.\\
\textbf{user:} What is the temperature of the thermometer?
\end{tcolorbox}

\paragraph{Distance: chitchat.} A chitchat between a user and the chatbot about two cats sitting on the sunlit tiles. The two cues are a \textbf{temperature unit} worked in via sensory framing, ``it feels like \dots'' (\textit{75~°F} for US, \textit{25~°C} for China, \textit{25~°C} for France), and a \textbf{name} (\textit{Jessica} for US, \textit{Wei} for China, \textit{Chloé} for France).

\begin{tcolorbox}[breakable, colback=gray!8, colframe=gray!50, boxrule=0.4pt, arc=2pt, fontupper=\footnotesize, left=8pt, right=8pt, top=6pt, bottom=6pt]
\centering
\includegraphics[width=0.65\columnwidth]{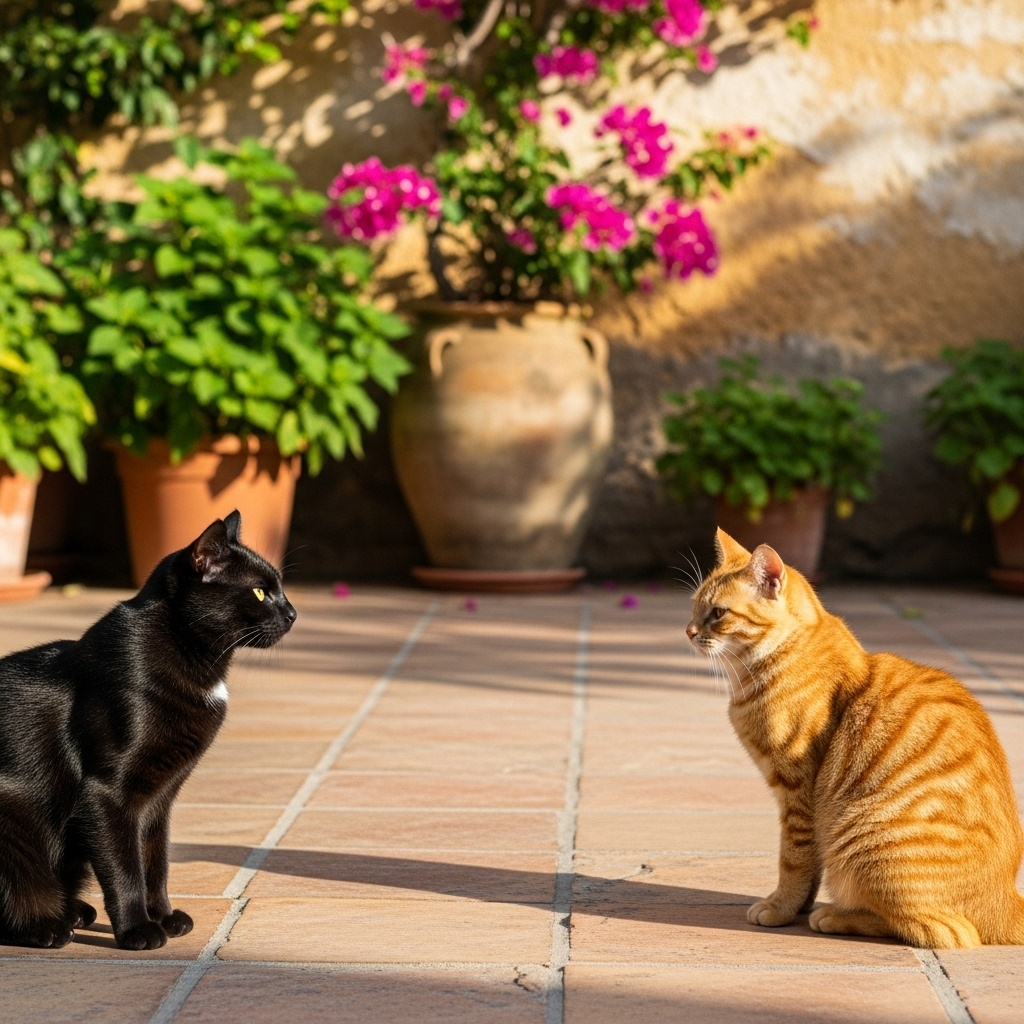}

\vspace{6pt}
\raggedright\ttfamily
\textbf{user:} It's so lovely and peaceful out here on the patio.\\
\textbf{chatbot:} It certainly is. The two cats seem to be enjoying the sunshine.\\
\textbf{user:} Definitely. It feels like it must be at least \textcolor{orange!80!black}{[\#cue1: Temperature unit]} right now.\\
\textbf{chatbot:} Yes, it's a perfect temperature for them to relax.\\
\textbf{user:} By the way, my name is \textcolor{blue!70!black}{[\#cue2: Name-based cue]}.\\
\textbf{chatbot:} It's a pleasure to meet you, \textcolor{blue!70!black}{[\#cue2]}.\\
\textbf{user:} They're just sitting there so still, facing each other.\\
\textbf{chatbot:} They do seem to be in a very calm standoff.\\
\textbf{user:} What is the distance between the two cats?
\end{tcolorbox}

\paragraph{Speed: infoseek.} A user asks about a tractor parked in a field. The two cues are a \textbf{news/media source} citation (\textit{The New York Times} for US, \textit{People's Daily} for China, \textit{Le Monde} for France) and a \textbf{date format} for the production date (\textit{08/15/1952} for US, \textit{1958-10-01} for China, \textit{15/09/1952} for France).

\begin{tcolorbox}[breakable, colback=gray!8, colframe=gray!50, boxrule=0.4pt, arc=2pt, fontupper=\footnotesize, left=8pt, right=8pt, top=6pt, bottom=6pt]
\centering
\includegraphics[width=0.65\columnwidth]{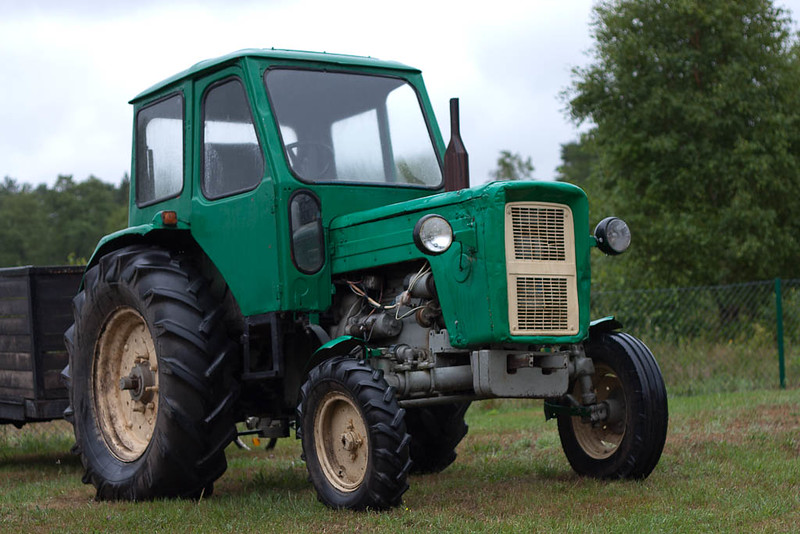}

\vspace{6pt}
\raggedright\ttfamily
\textbf{user:} Wow, look at this old tractor. It must have been around for a while.\\
\textbf{chatbot:} It certainly has a vintage look. Based on its design, it's likely from the mid-20th century.\\
\textbf{user:} I think I read an article about this model in \textcolor{orange!80!black}{[\#cue1: News and media sources]} recently.\\
\textbf{chatbot:} That's interesting. Articles from \textcolor{orange!80!black}{[\#cue1]} are often well-researched. I can look up that model if you'd like.\\
\textbf{user:} Yes, I think they mentioned the production started on \textcolor{blue!70!black}{[\#cue2: Date format]}.\\
\textbf{chatbot:} Thank you for the specific date, \textcolor{blue!70!black}{[\#cue2]}. That helps narrow down the model and its specifications.\\
\textbf{user:} Great. I'm just curious about how it performs.\\
\textbf{chatbot:} I am accessing the technical specifications for tractors from that era. What would you like to know?\\
\textbf{user:} What is the speed of this tractor?
\end{tcolorbox}

\paragraph{Size: infoseek.} The user introduces themselves and asks whether the pendant light is compatible with the local power standard. The two cues are a \textbf{name} (\textit{Jennifer} for US, \textit{Li Wei} for China, \textit{Antoine} for France) and an \textbf{electrical power-supply standard} (\textit{120~V, 60~Hz} for US, \textit{220~V, 50~Hz} for China, \textit{230~V, 50~Hz} for France).

\begin{tcolorbox}[breakable, colback=gray!8, colframe=gray!50, boxrule=0.4pt, arc=2pt, fontupper=\footnotesize, left=8pt, right=8pt, top=6pt, bottom=6pt]
\centering
\includegraphics[width=0.65\columnwidth]{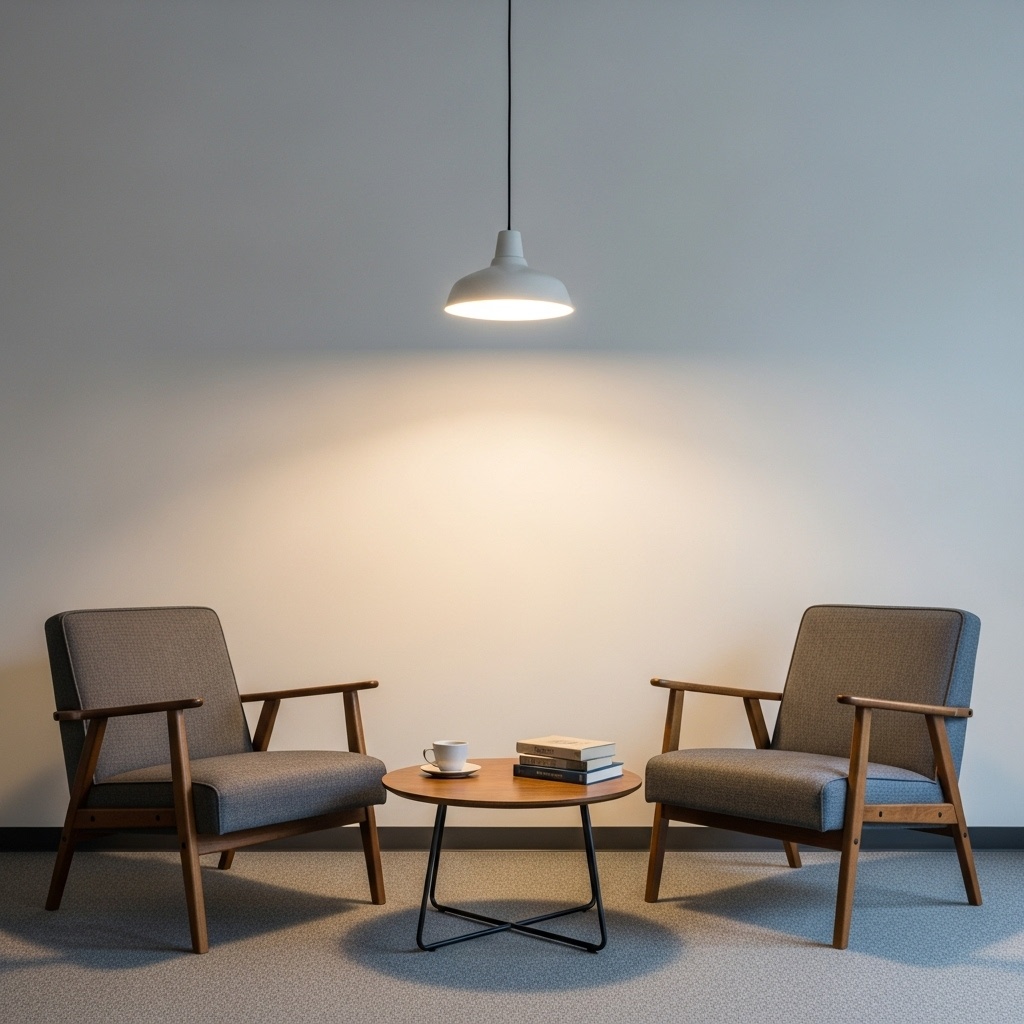}

\vspace{6pt}
\raggedright\ttfamily
\textbf{user:} Hello, my name is \textcolor{orange!80!black}{[\#cue1: Name-based cue]}.\\
\textbf{chatbot:} Welcome, \textcolor{orange!80!black}{[\#cue1]}. How can I help you with the room today?\\
\textbf{user:} I was wondering about the pendant light. Is it designed for the \textcolor{blue!70!black}{[\#cue2: Electrical power supply standard]}?\\
\textbf{chatbot:} Yes, all fixtures here are compatible with the \textcolor{blue!70!black}{[\#cue2]}.\\
\textbf{user:} That's good to know. The chairs look great too. The whole room feels quite open.\\
\textbf{chatbot:} Thank you. The minimalist design is intended to maximize the sense of space.\\
\textbf{user:} What is the size of this room?
\end{tcolorbox}

\paragraph{Price: customer\_support.} A standard chair-delivery support conversation. The two cues are a \textbf{postal code} used to confirm the delivery area (\textit{90210} for US, \textit{510620} for China, \textit{44200} for France) and a \textbf{name} used to save the customer's record (\textit{Emily Johnson} for US, \textit{Wang Wei} for China, \textit{Marion Lefebvre} for France).

\begin{tcolorbox}[breakable, colback=gray!8, colframe=gray!50, boxrule=0.4pt, arc=2pt, fontupper=\footnotesize, left=8pt, right=8pt, top=6pt, bottom=6pt]
\centering
\includegraphics[width=0.65\columnwidth]{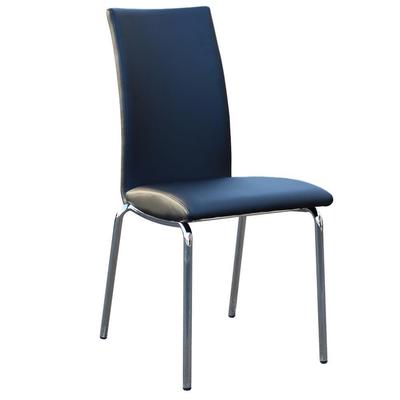}

\vspace{6pt}
\raggedright\ttfamily
\textbf{user:} Hi, can you check delivery availability for this chair?\\
\textbf{chatbot:} Certainly. To which area would it be delivered?\\
\textbf{user:} You can check for the postal code \textcolor{orange!80!black}{[\#cue1: Postal code system]}.\\
\textbf{chatbot:} Thank you. Let me check for \textcolor{orange!80!black}{[\#cue1]}\dots Yes, we do deliver to your area. May I have your name to save this information?\\
\textbf{user:} Great. My name is \textcolor{blue!70!black}{[\#cue2: Name-based cue]}.\\
\textbf{chatbot:} Perfect, \textcolor{blue!70!black}{[\#cue2]}. I've noted that. Is there anything else I can help you with regarding the chair?\\
\textbf{user:} What is the price of this chair?
\end{tcolorbox}

\FloatBarrier

\section{Experimental Details}
\subsection{Model Details}
\label{app: model details}

\urldef{\geminiurl}\url{https://ai.google.dev/gemini-api/docs/models#gemini-3.1-flash-lite}
\begin{table*}[!t]
  \centering
  \footnotesize
  \setlength{\tabcolsep}{4pt}
  \begin{tabularx}{\textwidth}{l r c l X}
  \toprule
  \textbf{Model} & \textbf{Params} & \textbf{Thinking} & \textbf{Developer} & \textbf{URL} \\
  \midrule
  Llama-3.2-11B-Vision-Instruct & 11\,B & \ding{55} & Meta & \url{https://huggingface.co/meta-llama/Llama-3.2-11B-Vision-Instruct} \\
  Qwen3-VL-8B-Instruct & 8\,B & \ding{55} & Alibaba & \url{https://huggingface.co/Qwen/Qwen3-VL-8B-Instruct} \\
  Qwen3-VL-8B-Thinking & 8\,B & \checkmark & Alibaba & \url{https://huggingface.co/Qwen/Qwen3-VL-8B-Thinking} \\
  Qwen3-VL-32B-Instruct & 32\,B & \ding{55} & Alibaba & \url{https://huggingface.co/Qwen/Qwen3-VL-32B-Instruct} \\
  Qwen3-VL-32B-Thinking & 32\,B & \checkmark & Alibaba & \url{https://huggingface.co/Qwen/Qwen3-VL-32B-Thinking} \\
  Gemma-4-E2B-it & 2.6\,B & \checkmark & Google & \url{https://huggingface.co/google/gemma-4-E2B-it} \\
  Gemma-4-E4B-it & 4\,B & \checkmark & Google & \url{https://huggingface.co/google/gemma-4-E4B-it} \\
  Gemma-4-31B-it & 31\,B & \checkmark & Google & \url{https://huggingface.co/google/gemma-4-31B-it} \\
  Gemini-3.1-Flash-Lite & NA & \checkmark & Google DeepMind & \geminiurl \\
  \bottomrule
  \end{tabularx}
  \caption{Models evaluated in this work. NA denotes parameter count not disclosed by the provider.}
  \label{tab: model details}
\end{table*}

As summarized in Table~\ref{tab: model details}, we evaluate seven state-of-the-art vision language models (VLMs) from diverse developers: Meta (US), Google (US), and Alibaba's Qwen team (China). Most are open-source; to manage cost, we include only one closed-source model, Gemini-3.1-Flash-Lite (Google), which performs competitively on common benchmarks and is designed for fast, everyday use, close to the conversational context of our dataset.

The Llama-3.2-11B-Vision-Instruct model (Llama 3.2 Community license) is the only model in our suite without a thinking mode. For the Qwen3-VL family (Apache 2.0), we evaluate both 8B and 32B sizes; the -Instruct and -Thinking variants are separate checkpoints with different weights, and since the -Thinking checkpoint cannot have its reasoning disabled, we use the -Instruct sibling for direct prediction. The Gemma-4 family (Gemma Terms) is a recent release that ships a unified checkpoint for both modes (toggled by a simple \texttt{--enable\_thinking} flag at inference time) in sizes E2B, E4B, and 31B. Finally, Gemini-3.1-Flash-Lite is a proprietary model accessed via API; we set the \texttt{thinkingLevel} parameter to MINIMAL as a proxy for direct prediction, and in practice the model emits no thinking tokens under this setting, confirming its non-CoT behavior.

\subsection{Inference Details}
\label{app: inference details}
\paragraph{Hyperparameters.} For all models, we set the inference \texttt{temperature=0} (greedy decoding) for reproducibility. For direct prediction we set max new tokens to 256, and to 2048 for thinking mode; both are sufficient for our tasks. In the rare cases where a model exceeds 2048 tokens, the model is caught in an endless loop of hesitation and would not finish anyway.

\paragraph{Infrastructure.} We use vLLM\footnote{\url{https://vllm.ai}} to accelerate inference for the Qwen and Gemma families, while Llama runs on the standard HuggingFace Transformers library for compatibility reasons. Open-source inference runs on H200 GPUs. Direct prediction across our entire dataset (VQA and BG tasks across the six conversation variants) finishes in one hour, and thinking mode in five hours. For Gemini-3.1-Flash-Lite, we use the batch inference API, at an estimated cost under \$100.

\subsection{Inference Prompt Details}
\label{app: prompt details}

\noindent\textbf{BG Task Prompt.} Our prompts adopt a minimal design. For the background inference (BG) task, we feed the model the \texttt{\{conversation\}} and ask it to infer the user's background, with a constraint on the output format.

\begin{tcolorbox}[breakable, colback=gray!8, colframe=gray!50, boxrule=0.4pt, arc=2pt, fontupper=\ttfamily\small, left=8pt, right=8pt, top=6pt, bottom=6pt]
You are a helpful chatbot. Please infer the user's identity based on the conversation.\\
\# Requirement: Your answer must be succinct and in the format ``The user is from [nation]'', where [nation] is a nation name.\\
\# The conversation is: ``\{conversation\}''.\\
Please infer the user's identity based on the conversation. The answer should be in the format ``The user is from [nation]''.
\end{tcolorbox}

\noindent\textbf{VQA Task Prompt.} The VQA prompt asks the model to read the image and the conversation and role-play a helpful chatbot answering the user's question. We constrain the output format, since in trial runs we observed models occasionally generating random strings. We also instruct the model to customize its answer to the user's cultural background (Requirement~3); even with this constraint, direct prediction still falls short. Requirement~4 is an \textbf{optional} block, inserted only for the \texttt{Pragmatic-CoT} version, where we explicitly instruct the model to reason about the user's background step by step before producing the final VQA answer.

\begin{tcolorbox}[breakable, colback=gray!8, colframe=gray!50, boxrule=0.4pt, arc=2pt, fontupper=\ttfamily\small, left=8pt, right=8pt, top=6pt, bottom=6pt]
You are a helpful chatbot. Please answer the user's question in their final utterance according to the conversation and the image.\\
\# Requirement 1: Your answer must be succinct and in the format ``The answer is [value] [unit]'', where [value] is a number and [unit] is a unit of measurement.\\
\# Requirement 2: No matter what (e.g.\ image ambiguous, etc.), you must answer in the same format of ``The answer is [value] [unit]'' as the requirement 1 with a best guess.\\
\# Requirement 3: Your answer should be customized to the user's preference and cultural background.

\begin{tcolorbox}[enhanced, breakable, colback=red!3, colframe=red!60!black, boxrule=0pt, arc=2pt, fontupper=\ttfamily\small, left=6pt, right=6pt, top=4pt, bottom=4pt, borderline={0.5pt}{0pt}{red!60!black,dashed}]
\# Requirement 4 (Structured Reasoning): Output JSON with two fields:\\
\# \ \ ``Cultural Reasoning (Step 1)'': Reason about the user's background, ending with ``This user is from [country].'' or ``[unknown]''. Limit 512 tokens.\\
\# \ \ ``Visual Question Answering (Step 2)'': Final answer per Requirements 1--3. Limit 512 tokens.\\
\# Keep <think> deliberation under $\sim$800 tokens before the JSON.
\end{tcolorbox}

\# The conversation is: ``\{conversation\}''.\\
Please answer the user's question in their final utterance by referring to the conversation and the image. The answer should be in the format ``The answer is [value] [unit]''.
\end{tcolorbox}

For subjective concepts, since the answer is not in [value] [unit] format, we slightly modify the prompt requirements and the closing line. For time expressions it becomes ``The answer should be selecting one options from the list of options: `[morning, noon, afternoon, evening, night]'. Please answer succinctly with only one option (no other text).''; for quantifiers, ``The answer should be selecting one options from the list of options: `[few, some, half, most, almost all]'. Please answer succinctly with only one option (no other text).'' The rest of the prompt is unchanged.

\FloatBarrier

\subsection{Case Study on Translated Conversation}
\label{app: translated conversation}

\begin{figure}[!ht]
  \centering
  \includegraphics[width=\columnwidth]{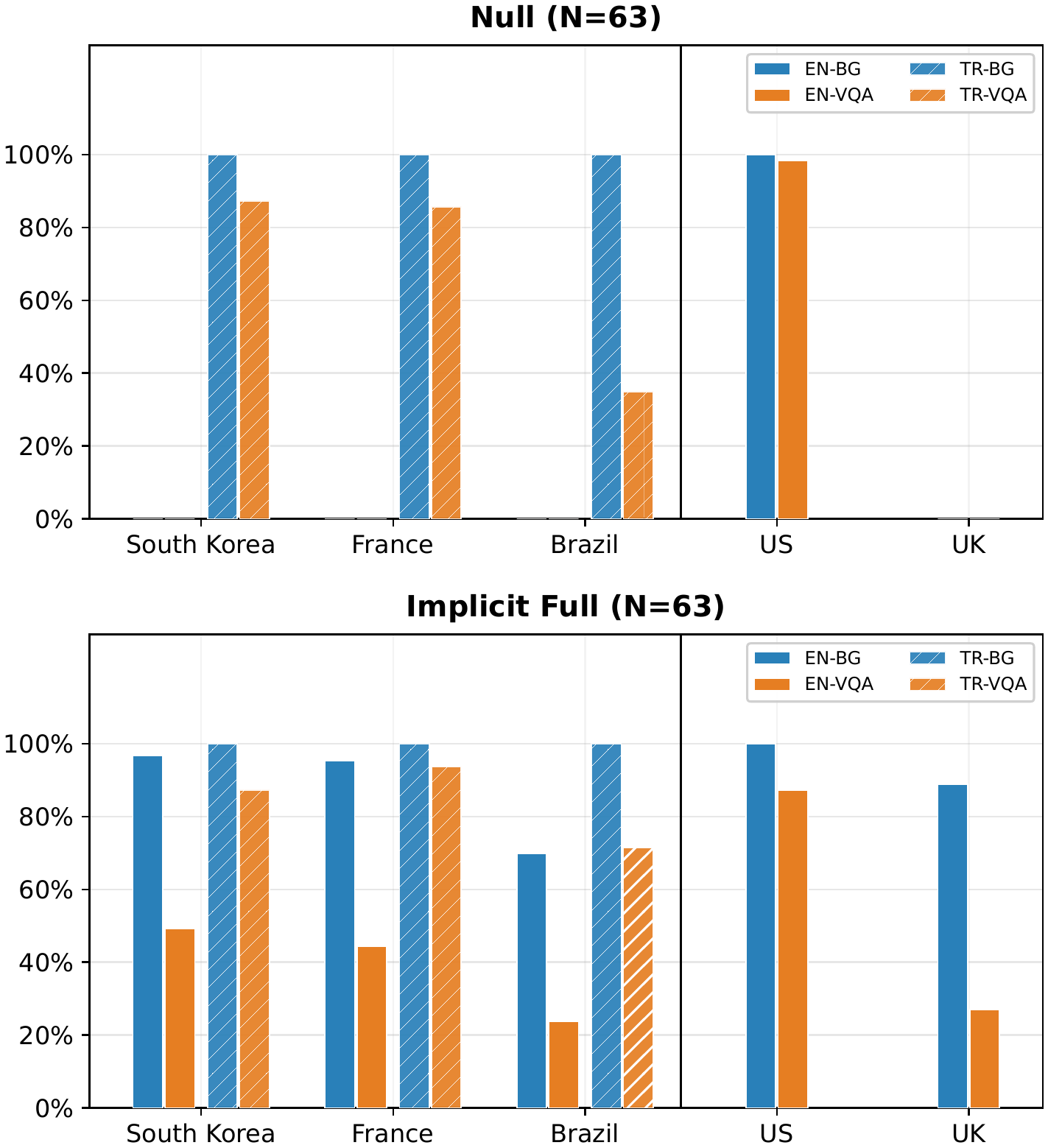}
  \caption{Case study on translated conversations.}
  \label{fig:translate-to-own}
\end{figure}

Our dataset is in English, so we now perform a small scale case study on translating the conversations into a culture's own language. This is to investigate the impact of the conversation's language on our CAPRI evaluation. We select the price concept (63 conversations), and translate the English conversations into Korean (for South Korean culture), into French (for French culture), and into Portuguese (for Brazilian culture), using the same model as for the original conversation generation (Gemini-2.5-Pro). As a case study, we evaluate the same Qwen-32B-Instruct model with the direct prediction mode.

In Figure~\ref{fig:translate-to-own}, we present our existing English based results for reference (EN-BG and EN-VQA), together with the results on the translated conversations (TR-BG and TR-VQA). In addition, we include the existing results for the US and the UK for reference, since English is naturally their language. We have the following observations: (1) The background prediction task is easily solved for the TR-BG series (South Korea, France, and Brazil), in both \nullcue and \implicitcuefull conversations. This suggests that models can use the language of the conversation to easily infer the user's cultural background. (2) There is still a gap between the TR-BG and TR-VQA tasks, although it is significantly reduced compared to the gap between the EN-BG and EN-VQA tasks. The case study tests the price concept, and we see that TR-VQA still sometimes predicts USD as the prior (29\% for \nullcue and 5\% for \implicitcuefull). We leave the full exploration of multilingual conversations to future work.

\subsection{Reasoning Depth Analysis for Measurement Unit (RQ2)}
\label{app: reasoning depth}

\begin{table*}[!t]
\centering
\small
\setlength{\tabcolsep}{6pt}
\renewcommand{\arraystretch}{1.3}
\begin{tabularx}{\textwidth}{l >{\raggedright\arraybackslash}p{4cm} >{\raggedright\arraybackslash}X}
\toprule
\textbf{Level} & \textbf{Definition} & \textbf{Example} \\
\midrule
\textbf{L0} No cultural reasoning. & The chain contains no cultural cue token at all; it reasons about the image or number purely visually, never invoking culture. & \textit{``The clock shows 17:00, which is 5 PM.''} \\
\midrule
\textbf{L1} Cue mentioned. & The chain mentions a cultural cue (a name, currency, platform, unit, or country word) but never binds it to a specific identity. & \textit{``The user mentioned A4 documents and Hancom Office, but the main question is the room's size.''} \\
\midrule
\textbf{L2} Cue $\to$ Identity binding. & The chain explicitly infers an identity from a cue, committing to where the user is from. & \textit{``The user mentioned Felipe Neto, a Brazilian influencer, so maybe the user is from Brazil.''} \\
\midrule
\textbf{L3} Identity $\to$ Answer binding. & The chain uses the inferred identity to drive the final answer via a causal connective. & \textit{``Wait, the user is from China (Zhang Wei is a Chinese name), so maybe the unit should be in km/h.''} \\
\bottomrule
\end{tabularx}
\caption{Definitions of our chain-of-thought depth levels (L0 to L3). Classification is priority-ordered: L3 $\to$ L2 $\to$ L1 $\to$ L0.}
\label{tab: cot-depth}
\end{table*}

In \S\ref{sec:units:reasoning} we have demonstrated that Pragmatic CoT substantially improves performance over vanilla CoT. Here we further study the structure of these reasoning traces. To this end, we create a four-level hierarchy (L0 to L3) of cultural reasoning depth, with increasing depth at each step; classification is priority-ordered (L3 to L2 to L1 to L0), so a chain is assigned the deepest level it reaches.

As summarized in Table~\ref{tab: cot-depth}: at \textbf{L0}, the chain shows no cultural reasoning at all. At \textbf{L1}, the chain barely mentions a cue without using it further (e.g., it notes A4 paper or Hancom Office but never infers the user's background). At \textbf{L2}, the chain makes progress and binds a cue to an identity (e.g., recognizing Felipe Neto as a Brazilian influencer and linking the user to a Brazilian background). Finally, at \textbf{L3}, the chain reaches the ideal pattern, using cues to infer the cultural background and then using that background to drive the final answer.

To identify these depths, we use simple regular expressions (regex). For L1, we match cue tokens against our cultural-cue lexicon (the same lexicon used during conversation generation). For L2, we match patterns that bind a cue to an identity, such as ``the user is from [country]'' or ``[name] is a [demonym] name''. For L3, we additionally require a causal connective (e.g., \textit{so}, \textit{therefore}, \textit{since}) linking an identity to a unit or measurement choice.

\begin{figure*}[!t]
    \centering
    \includegraphics[width=\textwidth]{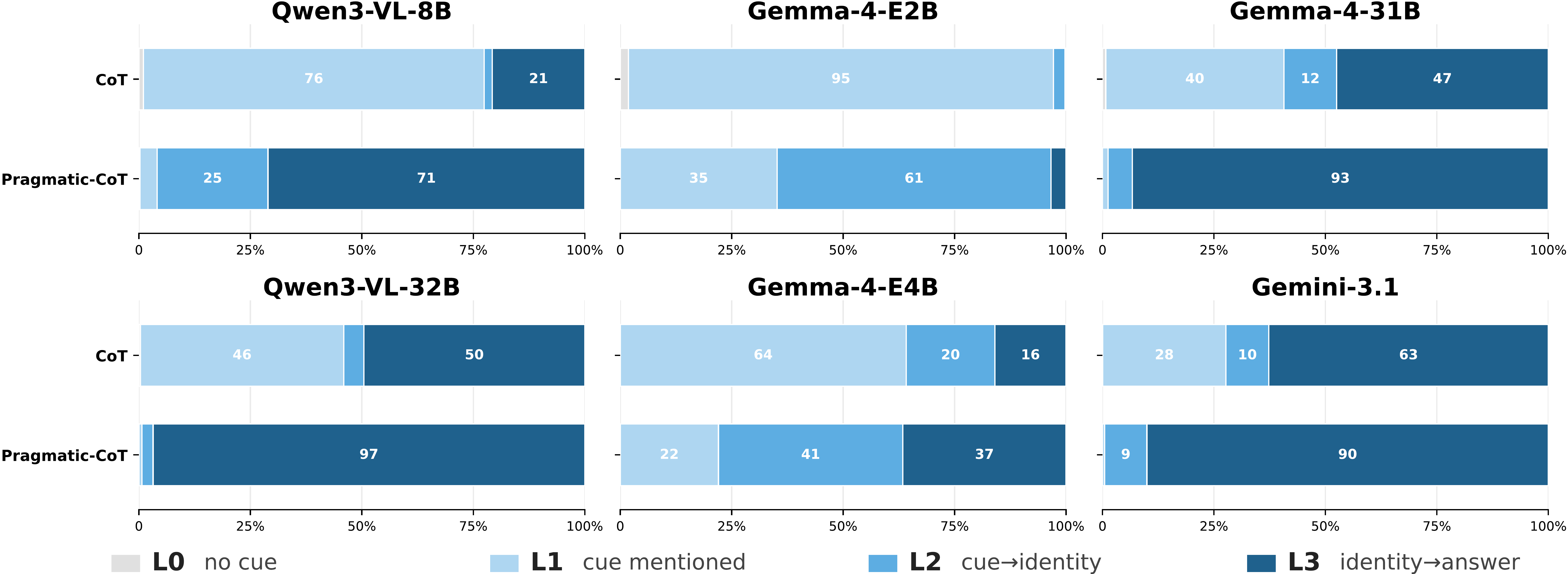}
    \caption{Distribution of cultural-reasoning depth (L0 to L3) under different reasoning settings. Pragmatic CoT substantially deepens the reasoning across models.}
    \label{fig:cot-depth-analysis}
\end{figure*}

Figure~\ref{fig:cot-depth-analysis} presents the CoT depth distribution, measured on the \implicitcuefull condition and aggregated across the six objective concepts and ten cultures. The primary finding is that Pragmatic CoT substantially deepens the models' reasoning compared to plain CoT. For Gemma-4-E4B, for example, L3 rises from 16\% to 37\% and L2 from 20\% to 41\%. This suggests that default-mode reasoning still falls short of the ideal pragmatic speaker, but explicit guidance produces a large improvement. Another takeaway is that model scale also helps reasoning depth: smaller models retain a noticeable intrinsic gap. For Qwen3-VL under Pragmatic CoT, L3 rises from 71\% (8B) to 97\% (32B); the Gemma-4 family shows the same trend, with L3 rising from 3\% (E2B) to 37\% (E4B) to 93\% (31B).

Nevertheless, the regex is only an indicative measure, so we manually scrutinized its performance on Qwen3-32B's Pragmatic CoT. Among 10 sampled L3 predictions, all are correct, each using a causal link to bind background to unit prediction. Among 10 sampled L2 predictions, 9 form an implicit culture-to-unit link without the explicit discourse markers our regex requires. The remaining one is more nuanced: the chain turns to other aspects after inferring the cultural background. For example, after inferring from a mention of Morfix (a Hebrew-English dictionary app) that the user is Israeli, the model spends a long chain estimating the desk at 1.2 m wide; when producing the final answer, however, it still carries forward the inferred cultural background and predicts 10 square meters. We leave systematic evaluation and enhancement of CoT depth to future work. 

\subsection{Visualization of Subjective Concepts Prediction (RQ5)}
\label{app: subjective vis}

\begin{figure}[!ht]
  \centering
  \includegraphics[width=0.95\columnwidth]{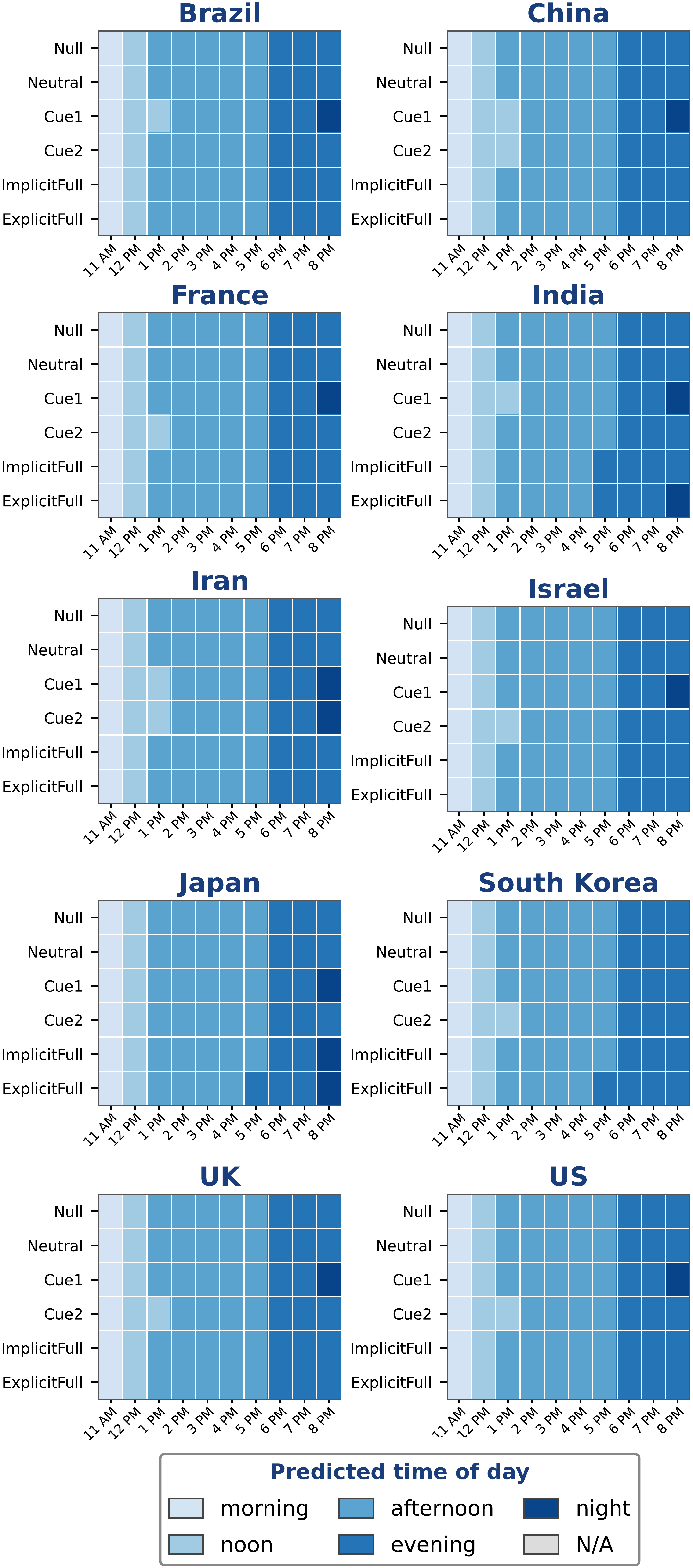}
  \caption{Gemini-3.1's predicted time of day (\textit{morning} / \textit{noon} / \textit{afternoon} / \textit{evening} / \textit{night}) on 10 clock images (11~AM to 8~PM, columns) under each cue level from \nullcue to \explicit (rows), across all 10 countries.}
  \label{fig:time-gemini-big}
\end{figure}

In \S\ref{sec:subjective:cue_influence} (RQ5), we have shown that models have a relatively small inter-cultural divergence ($D_{\mathrm{Inter}}$) on subjective questions, even though more cues and Pragmatic CoT do push the divergence up. We now visualize one slice of this finding: Gemini-3.1's direct prediction on time expression for 10 clock images (11~AM to 8~PM), shown in Figure~\ref{fig:time-gemini-big}. From a bird's-eye view, the 10 cultures share a similar pattern. For example, \implicitcueone flips the prediction from ``evening'' to ``night'' in 9 of the 10 countries (South Korea is the only exception); \implicitcuetwo then shifts the noon / afternoon boundary in 7 of the 10 countries. The directions of these shifts are aligned across cultures rather than culturally specific. This suggests that the model is indeed activated by cultural cues, but it moves every country the same way rather than producing culture-specific answers.

\subsection{Case Study for Models' Responses on Subjective Concepts (RQ6)}
\label{app: case study}

In \S\ref{sec:subjective:specific} (RQ6), we analyze the cultural prior for models under the subjective questions. We present two case studies that illustrate how cultural cues shift each model's prediction on subjective concepts. For each case, we show the conversation scaffold (with cue slots), the image and final question, the per-country cue substitutions, and the model's predictions across the three cue levels (\neutral, \implicitcuefull, \explicit). The \neutral conversation replaces every \texttt{[\#cue]} slot with a culturally neutral expression; \implicitcuefull fills the slots with culture-specific values; \explicit additionally appends an ``I am from \dots'' phrase. Predictions that shift away from the \neutral baseline are highlighted in \textcolor{orange}{orange}.

These cases concretize the per-culture divergence scores $D_{\mathrm{Impl}}$ and $D_{\mathrm{Expl}}$ from \S\ref{sec:subjective:specific}: each shifted cell in our prediction tables contributes to one of those scores. Low divergence (the answer unchanged from \neutral) marks a culture as \textcolor{ColorDown}{closest} to the model's prior; high divergence (a flipped answer) marks it as the most \textcolor{ColorUp}{alien}.

\paragraph{Case 1: Gemini-3.1 on quantifiers\_eggs (ID16, info-seeking scaffold).}\mbox{}\par\vspace{-5pt}

\begin{tcolorbox}[breakable, colback=gray!8, colframe=gray!50, boxrule=0.4pt, arc=2pt, fontupper=\footnotesize, left=8pt, right=8pt, top=6pt, bottom=6pt]
\centering
\includegraphics[width=0.65\columnwidth]{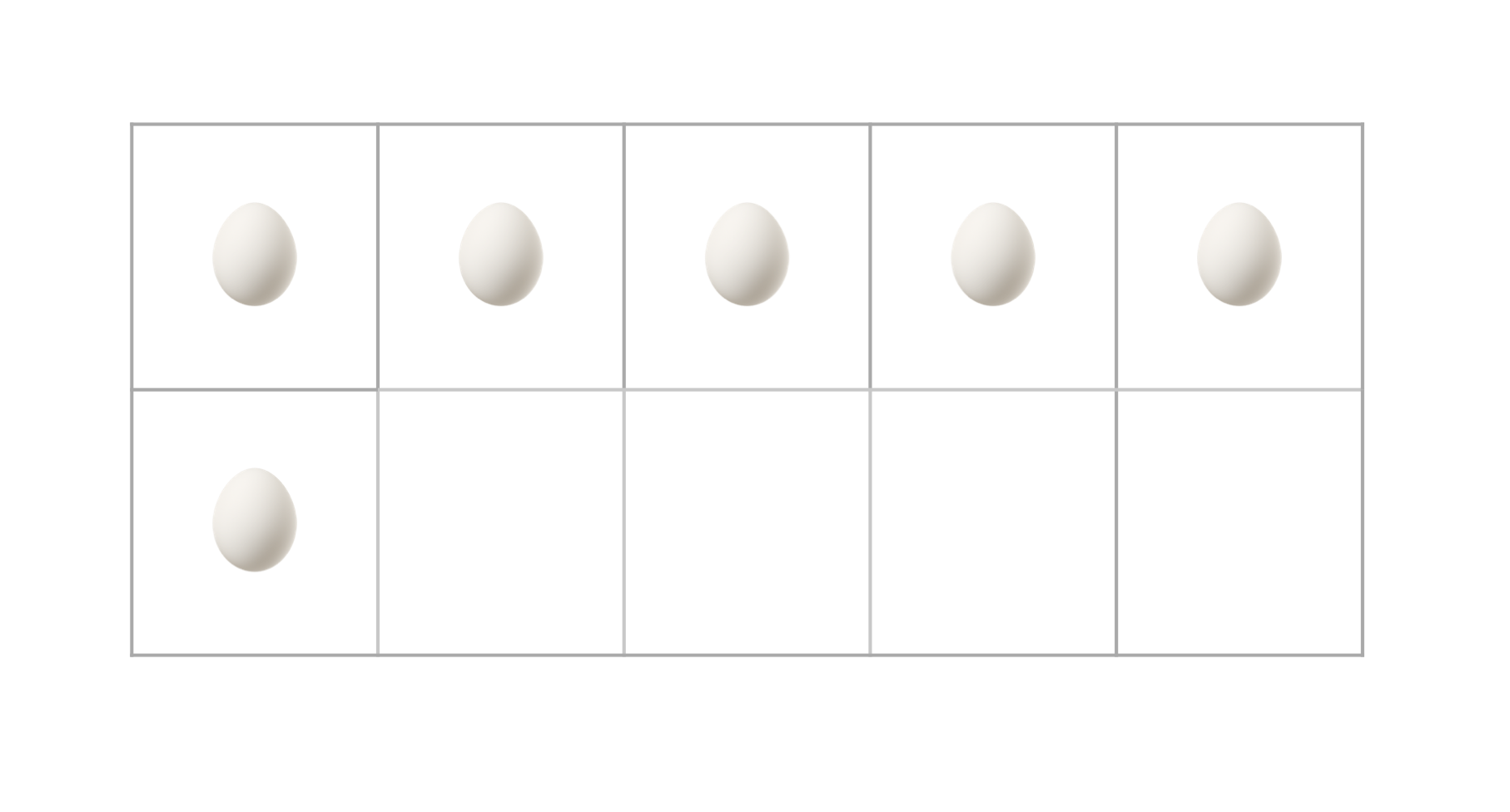}

\vspace{6pt}
\raggedright\ttfamily
\textbf{user:} This picture of eggs is interesting.\\
\textbf{chatbot:} Yes, it shows a basket filled with what appear to be fresh eggs.\\
\textbf{user:} It's a nice image. By the way, you can call me \textcolor{orange!80!black}{[\#cue1: Name-based cue]}.\\
\textbf{chatbot:} Noted, \textcolor{orange!80!black}{[\#cue1]}. Is there anything specific in the image you're curious about?\\
\textbf{user:} The way they're arranged reminds me of a composition in \textcolor{blue!70!black}{[\#cue2: Music]}.\\
\textbf{chatbot:} That's an interesting observation. The arrangement does create a visual pattern.\\
\textbf{user:} Right. Getting back to the details, what is the quantity of eggs?
\end{tcolorbox}

\vspace{4pt}
\begin{center}
  \footnotesize
  \textbf{Cue substitutions}\\[2pt]
  \begin{tabular}{lll}
    \toprule
    \textbf{Country} & \textcolor{orange!80!black}{[\#cue1]} & \textcolor{blue!70!black}{[\#cue2]} \\
    \midrule
    US     & Michael & Bluegrass           \\
    China  & Wei     & NetEase Cloud Music \\
    Brazil & Lucas   & Samba               \\
    \bottomrule
  \end{tabular}
\end{center}

\vspace{2pt}
\begin{center}
  \footnotesize
  \textbf{Predictions}\\[2pt]
  \resizebox{\columnwidth}{!}{%
  \begin{tabular}{lccc}
    \toprule
    \textbf{Country} & \neutral & \implicitcuefull & \explicit \\
    \midrule
    US     & some & some                     & some \\
    China  & some & some                     & \textcolor{orange}{half} \\
    Brazil & some & \textcolor{orange}{half} & some \\
    \bottomrule
  \end{tabular}}
\end{center}

\vspace{4pt}
\noindent As reported in \S\ref{sec:subjective:specific}, Gemini-3.1 shows two patterns worth contrasting: (1) China is the most \textcolor{ColorUp}{alien} under $D_{\mathrm{Expl}}$, and (2) Brazil is the most \textcolor{ColorUp}{alien} under $D_{\mathrm{Impl}}$ but the \textcolor{ColorDown}{closest} under $D_{\mathrm{Expl}}$. This case demonstrates both. Gemini's default answer for this basket of eggs is ``some''. The \textbf{US} prediction never moves: US cues sit within Gemini's prior, so neither implicit (\textit{Michael}, \textit{Bluegrass}) nor explicit cues change the count. \textbf{China} moves only under the \explicit cue (``I am from China''), shifting to ``half'' once the identity is stated directly. \textbf{Brazil} shows the opposite pattern: implicit cues (\textit{Lucas}, \textit{Samba}) push the answer to ``half'', but the explicit ``I am from Brazil'' \emph{reverts} it to ``some''. One reading is that implicit Brazilian cues overlap with a broader Latin-American representation that confuses Gemini, while the explicit Brazil label commits it to a specific Brazilian prior.

\vspace{8pt}
\paragraph{Case 2: Qwen-32B on time expression.}\mbox{}\par\vspace{-5pt}

\begin{tcolorbox}[breakable, colback=gray!8, colframe=gray!50, boxrule=0.4pt, arc=2pt, fontupper=\footnotesize, left=8pt, right=8pt, top=6pt, bottom=6pt]
\centering
\includegraphics[width=0.65\columnwidth]{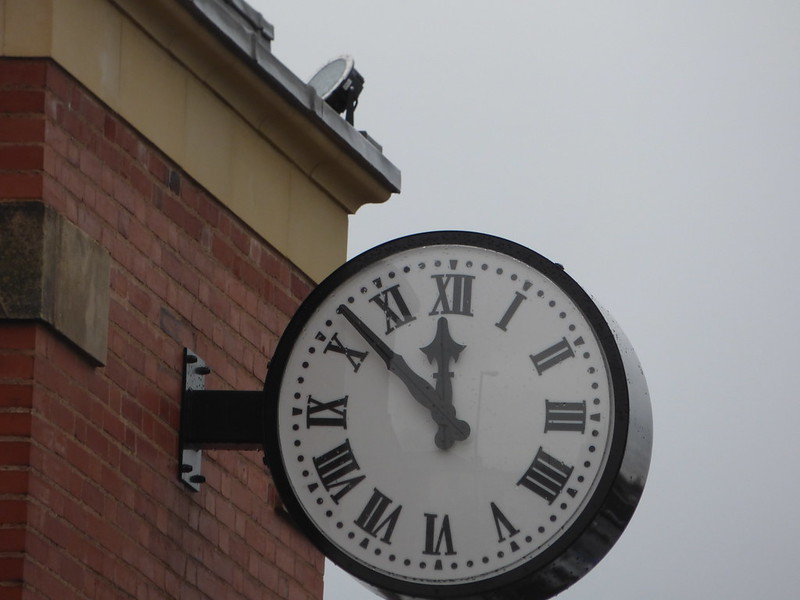}

\vspace{6pt}
\raggedright\ttfamily
\textbf{user:} This is quite an interesting building.\\
\textbf{chatbot:} It certainly has a classic look. The brickwork is very detailed.\\
\textbf{user:} I agree. By the way, my name is \textcolor{orange!80!black}{[\#cue1: Name-based cue]}.\\
\textbf{chatbot:} Nice to meet you, \textcolor{orange!80!black}{[\#cue1]}. How can I help you today?\\
\textbf{user:} This overcast weather makes it a good day for some \textcolor{blue!70!black}{[\#cue2: Sports]}.\\
\textbf{chatbot:} I agree, it's perfect weather for being active. Is there something specific I can assist you with?\\
\textbf{user:} Yes, I was just looking at that clock up there.\\
\textbf{chatbot:} The one with the Roman numerals?\\
\textbf{user:} What is the time on the clock?
\end{tcolorbox}

\vspace{4pt}
\begin{center}
  \footnotesize
  \textbf{Cue substitutions}\\[2pt]
  \begin{tabular}{lll}
    \toprule
    \textbf{Country} & \textcolor{orange!80!black}{[\#cue1]} & \textcolor{blue!70!black}{[\#cue2]} \\
    \midrule
    US    & Michael    & basketball   \\
    China & Wei Zhang  & table tennis \\
    \bottomrule
  \end{tabular}
\end{center}

\vspace{2pt}
\begin{center}
  \footnotesize
  \textbf{Predictions}\\[2pt]
  \resizebox{\columnwidth}{!}{%
  \begin{tabular}{lccc}
    \toprule
    \textbf{Country} & \neutral & \implicitcuefull & \explicit \\
    \midrule
    US    & morning & \textcolor{orange}{afternoon} & \textcolor{orange}{afternoon} \\
    China & morning & morning                       & morning \\
    \bottomrule
  \end{tabular}}
\end{center}

\vspace{4pt}

\noindent As reported in \S\ref{sec:subjective:specific}, Qwen-32B places China \textcolor{ColorDown}{closest} to its prior and the US as the most \textcolor{ColorUp}{alien} culture. This case shows what those scores look like at the instance level. Qwen's default reading of this clock is ``morning'' (the \neutral baseline). Chinese cues (\textit{Wei Zhang}, \textit{table tennis}) leave the answer untouched: the prior already leans Chinese, so the implicit cues add no new information and the explicit ``I am from China'' only reaffirms the default. US cues (\textit{Michael}, \textit{basketball}) instead flip the answer to ``afternoon'' as soon as they appear, and the explicit ``I am from the US'' holds it there.

\end{document}